\pdfoutput=1
\documentclass[11pt]{article}

\usepackage{ACL}

\usepackage{times}
\usepackage{latexsym}

\usepackage[T1]{fontenc}
\usepackage[utf8]{inputenc}

\usepackage{microtype}

\usepackage{inconsolata}

\usepackage{amsmath}
\usepackage{amsfonts}
\usepackage{graphicx}
\usepackage{subcaption}
\usepackage{tikz}
\usepackage{multirow}
\usepackage{float}
\usepackage{booktabs}
\usepackage{tcolorbox}
\usepackage{float}
\usepackage{multicol}
\usepackage{algorithm}
\usepackage{algpseudocode}
\usepackage{soul}
\usepackage{arydshln}
\usepackage{bbm}

\title{Graph-Induced Syntactic-Semantic Spaces in Transformer-Based \\ Variational AutoEncoders}

\author{Yingji Zhang$^{1\dagger}$,~ Marco Valentino$^{2}$, ~ Danilo S. Carvalho$^{1,3}$,\\ \textbf{~ Ian Pratt-Hartmann$^{1}$,~ Andr\'{e} Freitas$^{1,2,3}$} \\
  $^{1}$ Department of Computer Science, University of Manchester, United Kingdom\\
  $^{2}$ Idiap Research Institute, Switzerland\\
  $^{3}$ Cancer Biomarker Centre, CRUK Manchester Institute, United Kingdom\\
  \texttt{$^{1}$\{firstname.lastname\}@[postgrad.]$^{\dagger}$manchester.ac.uk}
  \\ \texttt{$^{2}$\{firstname.lastname\}@idiap.ch}}

\begin{document}
\maketitle
\begin{abstract}
The injection of syntactic information in Variational AutoEncoders (VAEs) has been shown to result in an overall improvement of performances and generalisation. An effective strategy to achieve such a goal is to separate the encoding of distributional semantic features and syntactic structures into heterogeneous latent spaces via multi-task learning or dual encoder architectures. However, existing works employing such techniques are limited to LSTM-based VAEs. In this paper, we investigate latent space separation methods for structural syntactic injection in Transformer-based VAE architectures (i.e., Optimus). Specifically, we explore how syntactic structures can be leveraged in the encoding stage through the integration of graph-based and sequential models, and how multiple, specialised latent representations can be injected into the decoder's attention mechanism via low-rank operators. Our empirical evaluation, carried out on natural language sentences and mathematical expressions, reveals that the proposed end-to-end VAE architecture can result in a better overall organisation of the latent space, alleviating the information loss occurring in standard VAE setups, resulting in enhanced performances on language modelling and downstream generation tasks.
% Moreover, we found that the induced latent separation and subsequent injection can enhance performances on language modelling and downstream generation tasks.
\end{abstract}

% The explicit representation of syntactic information in Variational AutoEncoders (VAEs) has been shown to improve performances on several language generation tasks, allowing for a more fine-grained structural organisation of the latent space. However, existing works to inject and disentangle syntactic features in VAEs are mostly limited to LSTM-based architectures, with techniques to achieve similar properties in larger language models being still under-explored. In this work, we focus on the explicit representation of syntactic structures in the latent space of Transformer-based VAEs (i.e., Optimus). Specifically, We show how structural syntactic information can be explicitly leveraged and disentangled in the encoding stage through the integration of graph-based and sequential encoders, and how the resulting representations can effectively guide the decoding phase via the attention mechanism. Our empirical evaluation reveals that the proposed end-to-end architecture can result in a better overall organisation of the latent space, alleviating the information loss occurring in standard VAE setups. Moreover, we found that the induced latent separation can enhance reconstruction and downstream performances on both natural language and mathematical inference tasks.

\section{Introduction}
% - The interpretation of statements is closely associated with the linguistic concept of compositionality, which integrates the relationship between of syntax and semantics. 
% - At the center of this notion (e.g. in Montague semantics) is the fact that words within a sentence combine following the syntactic structure.
% - Ideally, this relationship between syntax and semantics ideally would be observable and controllable within latent spaces.
% - However, despite the prevalence and success of contemporary LMs to ..., attempts to connect back to desirable formal properties have been timid.
% - This work focuses on .(exactly this problem).. where a systematic analysis ...
% - In order to guarantee a controlled setting for investigating the separation of syntax and semantics ...

Injecting explicit syntactic information in Variational AutoEncoders (VAEs) \cite{kingma2013auto} has led to improved performance on several language generation tasks, such as paraphrasing and translation \cite{dai2018syntax,chen-etal-2017-improved,felhi-etal-2022-exploiting,yang-etal-2021-syntactically}. Among existing techniques, a line of research explores syntactic injection via sentence-level semantics-syntax disentanglement, which consists in the explicit separation of distributional semantic and structural syntactic features through the optimisation of heterogeneous latent spaces \cite{bao-etal-2019-generating, chen-etal-2019-multi, zhang-etal-2019-syntax-infused}. Such methods, implemented under multi-task learning or dual encoder architectures, have been demonstrated to improve: (i) generation controllability and interpretability \cite{bao-etal-2019-generating,zhang2022quasi}, (ii) robustness and generalisation, and (iii) fine-grained representation and latent space organisation \cite{chen-etal-2019-multi}. However, most of these methods focus on LSTM-based VAEs, and their effectiveness for larger architectures based on Transformers, such as Optimus \citet{li2020optimus}), is still under-explored.

To combine the benefits of larger pre-trained VAEs and latent separation methods, this paper focuses on the injection of structural syntactic information in Transformer-based VAEs (i.e., Optimus \cite{li2020optimus}. 
Specifically, we investigate a first overarching research question: \textit{``RQ1: How can we best capture explicit syntactic information in the latent space of Transformer-based VAEs?''} 
we address this question by directly intervening on the Optimus architecture to induce a latent space separation via graph-based \cite{kipf2016semi} and sequential neural encoders \cite{devlin2018bert}. Specifically, our hypothesis is that Graph Neural Networks (GNNs) \cite{kipf2016semi,hamilton2017inductive,yun2020graph} can induce specialised and complementary latent representations that can better capture structural syntactic relations and alleviate the information bottleneck in VAEs' semantic encoder \cite{alemi2016deep,tenney2019you} (i.e. trade-off between semantics and syntax). 
%Our experimental results reveal that, while the Optimus setup adopting a BERT encoder struggles to capture syntactic structures in the latent space, a dual encoder configuration integrating BERT \cite{devlin2018bert} and Graph Neural Networks (GNNs) \cite{kipf2016semi,hamilton2017inductive,yun2020graph} can better support such property.

Subsequently, we focus on the problem of leveraging multiple, specialised latent spaces derived from the dual encoder architecture for decoding. This leads to several challenges (Figure \ref{fig:latent_space_geometry}) since (i) the syntactic representations may not possess a one-to-one mapping with the semantic representations (i.e., one syntactic structure can correspond to multiple sentence representations), (ii) the optimisation of heterogeneous latent spaces can result in different latent distributions, a feature that can affect decoding and language generation performance (iii) compared with an LSTM decoder, Transformer-based decoders (e.g., GPT2) are typically larger and contain information acquired during pre-training, being more difficult to control. 

Those challenges lead to our second research question: \textit{``RQ2. How can multiple, specialised latent spaces be effectively injected into the VAE decoder?''}
To answer this question, we investigate injection mechanisms for Transformer-based VAEs via the following methods: (i) we separately inject syntax and semantic representations into the attention weights of the decoder (i.e., Query and Key-Value), and (ii) consider low-rank injections, including \textit{addition}, \textit{memory} \cite{li2020optimus}, and \textit{tensor fusion} \cite{liu-etal-2018-efficient-low}, which directly operate over the attention weight matrices as low-rank operation can reduce information redundancy \cite{hu-etal-2022-fuse}. 

% We hypothesise that low-rank operations can help reduce information redundancy and better capture diverse and specialised features in the latent representations for guiding the generation process.

We perform extensive experiments to evaluate the resulting VAE architectures on both mathematical expressions \cite{valentino2023multioperational,meadows2023symbolic} and natural language explanatory sentences \cite{jansen2018worldtree}. Overall, our contributions can be summarised as follows:

\textbf{1.} We propose a dual encoder architecture for Transformer-based VAEs integrating graph-based and sequential models to better capture and disentangle semantic and structural syntactic features in multiple, specialised latent spaces. 

\textbf{2.}We explore the injection of such representations into the decoder of Transformer-based VAEs via low-rank vector operations to better guide the generation process.

\textbf{3.} We perform extensive experiments showing that the adoption of a graph-based encoder coupled with a transformer encoder can reduce the loss of information in the sentence bottleneck, resulting in improved reconstruction and language modelling. Overall, We found that the proposed VAE architecture can significantly improve performance and generalisation when compared to sentence-level VAE baselines.
% and existing VAE models in the literature.

Our complete experimental code is available at \footnote{\url{https://github.com/SnowYJ/sem_syn_separation}} to encourage future work in the field.

\section{Preliminaries} \label{sec:pre}
\begin{figure}[t]
    \centering
    \includegraphics[width=\columnwidth]{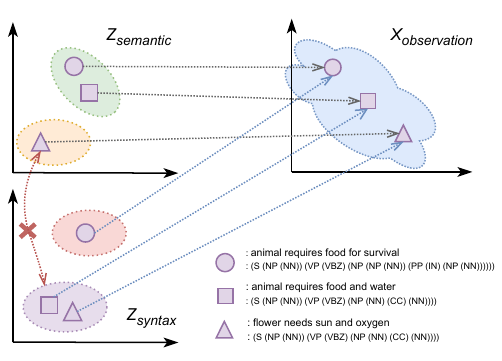}
    \caption{Decoding under heterogeneous syntactic-semantic spaces can result in two main challenges: (i) The syntactic representations may not possess a one-to-one mapping with the semantic representations (i.e., one syntactic structure can correspond to multiple sentence representations), (ii) the optimisation of heterogeneous latent spaces can result in different latent distributions, making generation hard to control.}
    \label{fig:latent_space_geometry}
\end{figure}

\paragraph{Latent Space Injection.} \label{sec:lsg} 
In Optimus, the transformation between latent (i.e., Gaussian) and observed (i.e., generated sentences) spaces can be done by intervening on the Key-Value attention weights of the decoder (i.e., GPT2) via \textit{memory} injection \cite{li2020optimus}. Specifically, the latent representation $z$ produced by the encoder (i.e., Bert) is concatenated into the original Key-Value weights of GPT2 as follows:
\[
\begin{aligned}
    \text{Attention}(Q, K, V) = \text{softmax}( \frac{Q [z;K]^T}{\sqrt{d}})[z;V]
\end{aligned}
\]
Where $Q$ has dimension $\mathbb{R}^{64 \times \text{seq}}$, and $[z;K], [z;V]$ have dimension $\mathbb{R}^{ 64 \times (\text{seq}+1)}$ (where 64 is the dimension of GPT2 attention, $\text{seq}$ is sequence length). %Since $Q$ represents the target, $K$ and $V$ represent the latent representations. 
In order words, the decoder model is explicitly guided in the generation process by conditioning $KV$ on $z$. In this work, however, we focus on heterogeneous representations encoding distributional semantic and structural syntactic features in separate latent spaces (see Figure \ref{fig:latent_space_geometry}). Such a separation requires going beyond the \textit{memory} injection setup and developing different methods to effectively condition the decoding process in GPT2.

\begin{figure*}[t]
    \centering
    \includegraphics[width=\textwidth]{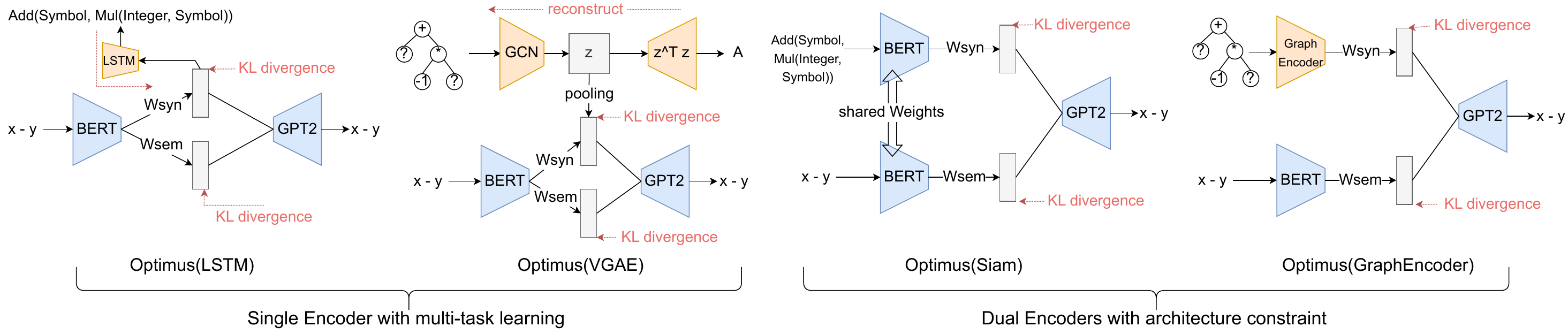}
    \caption{Overview of different methods to explicitly represent and disentangle syntactic information in the latent space of Transformer-based VAEs.}
    \label{fig:sem_syn_baselines}
\end{figure*}

\paragraph{Semantic-Syntax Relation.} \label{sec:shared} Following the \textit{principle of compositionality}, the semantics of a sentence can be seen as a composition of word-level semantics, induced by the meaning of individual words and their relations \cite{dowty2012introduction, yao-etal-2023-words}. Instead of considering sentence-level semantics only as a composition of word content as done in previous work on latent separation for LSTM-based VAEs \cite{bao-etal-2019-generating}, this work uses the notion of sentence semantics as word content plus positional elements (i.e. \textit{word order typology} \cite{sankaravelayuthanword}), which has been well captured by Transformer-based encoders (e.g, BERT). Under this constraint, mutual information naturally exists between semantics and syntax. Therefore, although separating semantic and syntactic features in heterogeneous latent spaces can lead to representations that are not geometrically aligned in the Gaussian space (Figure \ref{fig:latent_space_geometry}), such mutual information can be captured through low-rank injection \cite{zhang-etal-2019-syntax-infused}. Therefore, to guide the decoder, we investigate low-rank operators  $\otimes$, which directly work on QKV instead of token embeddings or the last hidden representation \cite{hu-etal-2022-fuse}.

\section{Methodology}
% Previous works have shown that explicitly injecting and disentangling syntactic information in the latent space of VAEs can improve the overall performances \cite{bao-etal-2019-generating,zhang-etal-2019-syntax-infused,huang-chang-2021-generating}. In this paper, we build upon this line of research, investigating how to induce such properties in Transformer-based models -- i.e, Optimus \cite{li2020optimus}.
In this paper, we build upon the line of research \cite{bao-etal-2019-generating,zhang-etal-2019-syntax-infused,huang-chang-2021-generating}, investigating how to induce semantic-syntax separation in Transformer-based models -- i.e, Optimus \cite{li2020optimus}.

Our methodology consists of two main phases. First, we investigate different encoding strategies to explicitly capture syntactic and structural information in a separate latent space. Subsequently, we explore techniques to fuse syntactic and semantic features and inject them into the decoder model.
Regarding the encoding phase, we explore four architectures based on two different configurations (i.e., \emph{multi-task learning} and \emph{dual encoder}) integrating both \emph{sequential} and \emph{graph-based} models under Optimus (Bert-GPT2) \textit{memory} setup 
 (see Figure \ref{fig:sem_syn_baselines}).
 Regarding the decoding phase, we consider the best encoding configuration in terms of syntactic representation and propose different injection mechanisms via low-rank operations over attention-weight matrices of GPT2. 

 The following sections describe each phase in detail (Sections \ref{sec:encoding_phase} and \ref{sec:decoding_phase}), including how the encoding and decoding stages are integrated into an end-to-end VAE architecture (Section \ref{sec:arc}).

\begin{figure*}[t]
    \centering
    \includegraphics[width=\textwidth]{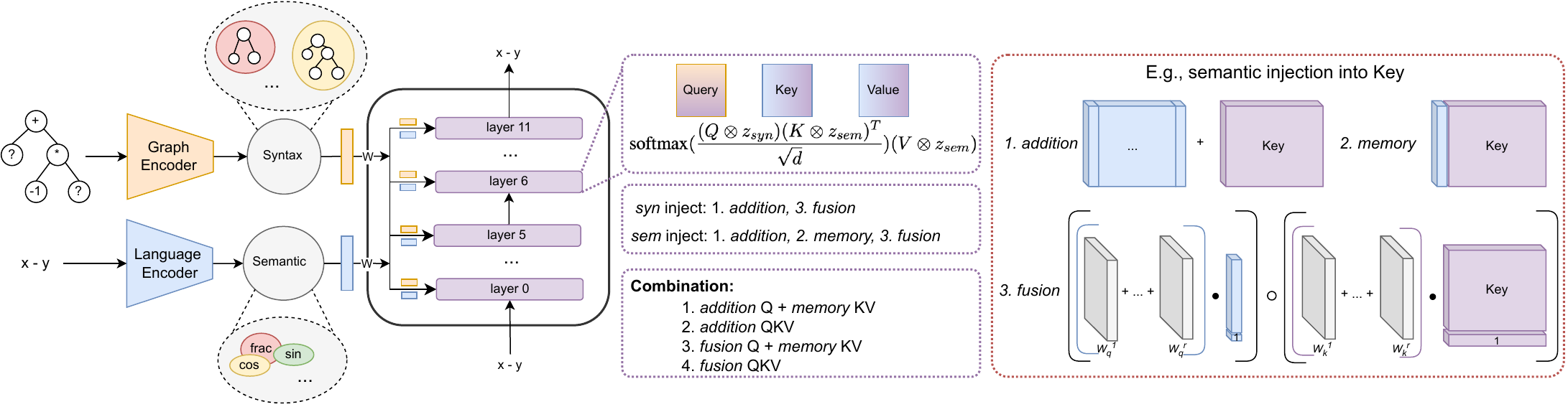}
    \caption{Architectural overview. Semantic and syntactic features are encoded into heterogeneous latent spaces via graph-based and sequential encoders. The resulting latent spaces are then injected into the GPT2 decoder via low-rank operations.}
    \label{fig:overview}
\end{figure*}
\subsection{Encoding Syntactic-Semantic Spaces}
\label{sec:encoding_phase}

\paragraph{Multi-Task Learning.}
\citet{bao-etal-2019-generating} proposed a multi-task learning strategy to achieve such a goal in LSTM-based VAEs via learning and fusing two distinct latent representations. They adopt a separate space for encoding explicit syntactic dependencies through the adoption of an LSTM decoder used to reconstruct flattened constituency parse trees. Here, we build upon this setup to enrich the latent representation in Optimus \cite{li2020optimus}. Specifically, given a separate latent syntax representation, $z_{syn}$, encoded via BERT \cite{devlin2018bert}, we explore the following mechanisms (see Figure \ref{fig:sem_syn_baselines}):
% \begin{enumerate}

\textbf{1.} Similarly to \cite{bao2019generating}, we adopt an LSTM \cite{10.1162/neco.1997.9.8.1735} decoder to generate linearised syntactic trees, where $z_{syn}$ is fed into the first hidden state of the LSTM. We refer to this configuration as \emph{Optimus (LSTM)}.
\textbf{2.} We jointly train a Variational Graph AutoEncoder (VGAE, \citet{kipf2016variational}) on syntactic trees, where the latent node embeddings are mean-pooled into a sentence-level syntax representation $z^{gcn}_{syn}$.  We refer to this configuration as \emph{Optimus (VGAE)}.
Here, the syntactic representations $z^{gcn}_{syn}$ and $z_{syn}$ can be optimized via MSE in a multi-task setting. Specifically, the general objective function can be formalised as:
\[
\begin{aligned}
& \mathcal{L}_{\text{VAE}} = \mathbb{E}_{q_\phi(z_{sem},z_{syn}|x)} \Big[ \log p_{\theta}( x | z_{sem}, z_{syn}) \Big] \\
& - \text{KL}(\phi(z_{sem}|x)||p(z)) - \text{KL}(\phi(z_{syn}|x)||p(z)) \\
& + \mathcal{L}_{\text{syn}}(z_{syn})
\end{aligned}
\]
% + \underbrace{\sum_{w \in vocab} t_w \log p(w|z_{sem})}_{word(BoW)} $t_w$ represents the ground-truth distribution of word $w$.
Where $q_\phi, p_{\theta}$ represent the encoder and decoder. The objective functions for optimising the syntactic spaces $\mathcal{L}_{\text{syn}}(z_{syn})$ can be specialised according to the model:
\[
\begin{aligned}
    \mathcal{L}^{lstm}_{\text{syn}}(z_{syn})&=\sum^n_{i=1} \log p(s_i|s_1, \dots, s_{i-1}, z_{syn}) \\
    \mathcal{L}^{vgae}_{\text{syn}}(z_{syn})&=\sum^{dim}_{j=1}(z^{j}_{gcn} - z_{syn}^j)^2 + \mathcal{L}^{vgae}(A, N)
\end{aligned}
\]
Where $s_i$ represents the token of a flattened syntax tree, while $A$ and $N$ are the Adjacent matrix and Node embeddings of the syntax tree. Additional details for the VGAE model and the optimisation of $\mathcal{L}^{vgae}$ can be found in the original paper \cite{kipf2016variational}. 

\paragraph{Dual Encoder.} In addition to the multi-task learning setup, we build upon \citet{zhang-etal-2019-syntax-infused,huang-chang-2021-generating} which propose two distinct language encoders to induce syntactic disentanglement. Specifically, we experiment with: 

\textbf{1.} Two distinct BERT encoders via a Siamese neural network. We refer to this configuration as \emph{Optimus (Siam)}.
\textbf{2.} A Graph encoder, such as GCN \cite{kipf2016semi}, GraphSAGE \cite{hamilton2017inductive}, and Graph Transformer (TransCONV, \citet{yun2020graph}), coupled with a BERT encoder. We refer to this configuration as \emph{Optimus (GraphEncoder)}.
Here, the general objective function can be formalised as:
\[
\begin{aligned}
& \mathbb{E}_{q^{sem}_\phi(z_{sem}|x), q^{syn}_\phi(z_{syn}|x_{syn})} \Big[ \log p_{\theta}( x | z_{sem}, z_{syn}) \Big]  \\
& - \text{KL}(\phi(z_{sem}|x)||p(z)) - \text{KL}(\phi(z_{syn}|x)||p(z))
\end{aligned}
\]
% & + \underbrace{\sum_{w \in vocab} t_w \log p(w|z_{sem})}_{word(BoW)}
Where $q^{sem}_\phi, q^{syn}_\phi$ represent semantic and syntax encoders respectively, while $x_{syn}$ represents the input for the syntax encoder. For graph encoders, we represent $x_{syn}$ using an adjacency matrix and node embedding pairs. For the language syntax encoder, on the other side,  we represent $x_{syn}$ as a flattened syntactic tree without word content.

As our experiments revealed that the \emph{dual graph-sequential encoder} configuration (i.e., \emph{Optimus (GraphEncoder)}) can achieve the best results in terms of syntactic representation (see Table \ref{tab:enoding_syntax}), we consider this setup for integration into an end-to-end VAE architecture (see Section \ref{sec:arc}).

\subsection{Decoding Heterogeneous Representations}
\label{sec:decoding_phase}
To preserve the separation of the latent spaces and, at the same time, leverage heterogeneous representations during decoding, we explore methods to inject semantic (i.e., $z_{sem}$) and syntactic space (i.e., $z_{syn}$) directly into the attention mechanism of GPT2 (via Query-Key-Value(QKV)). Specifically, we inject different latent representations to different attention weights:
\begin{equation}
\begin{split}
    \text{softmax}( \frac{(Q \otimes z_{syn}) (K \otimes z_{sem})^T}{\sqrt{d}})(V \otimes z_{sem}) \nonumber
\end{split}
\end{equation}
% & \text{Attention}(Q, K, V) = \\ \nonumber
Where $\otimes$ represents the latent injection operation.
As for syntactic injection ($z_{syn}$), we consider two kinds of low-rank operations $\otimes$, \textit{addition}, and \textit{fusion} \cite{liu-etal-2018-efficient-low}, which directly work on attention weights. As for \textit{addition}, we inject $z_{syn}$ into each low-rank token representation in Q, which can be formalised as follows:
\[
\begin{aligned}
    \tilde{Q} &= \sum_{i=1}^{seq} Q[i, :] + z_{syn}
\end{aligned}
\]
Where $\tilde{Q}$ represents the new Q values obtained after syntax injection. As for \textit{fusion}, we adapt the tensor fuse operation \cite{liu-etal-2018-efficient-low, hu-etal-2022-fuse}. In more detail, given a hyper-parameter, rank $r=4$, the $\tilde{Q}$ can be described as:
\[
% \left\{
\begin{aligned}
    \tilde{Q} &= (\sum_{i=1}^r W_q^i [Q;\mathbbm{1}]) \circ (\sum_{i=1}^r W_z^{i, syn} [z_{syn};\mathbbm{1}])
    % \tilde{K} &= (\sum_{i=1}^r W_{kv} [K;\mathbbm{1}]) \circ (\sum_{i=1}^r W_z^{i, sem} [z_{sem};\mathbbm{1}]) \\
    % \tilde{V} &= (\sum_{i=1}^r W_{kv} [V;\mathbbm{1}]) \circ (\sum_{i=1}^r W_z^{i, sem} [z_{sem};\mathbbm{1}])
\end{aligned}
% \right.
\]
Where $\mathbbm{1}$ is the matrix of ones, $W_z^{i, syn}$ and $W_q$ are the trainable linear transformations. 

As for semantic injection ($z_{sem}$), we consider three operations: \textit{addition}, \textit{memory}, and \textit{fusion}, where \textit{addition} and \textit{fusion} operations are the same as before but works on KV. \textit{Memory} is the same as Optimus \textit{memory} injection \cite{li2020optimus} as we described in section \ref{sec:pre}. 
% As for \textit{fusion}, the transformation can be formalised as follows:
% \[
% % \left\{
% \begin{aligned}
%     % \tilde{Q} &= (\sum_{i=1}^r W_q^i [Q;\mathbbm{1}]) \circ (\sum_{i=1}^r W_z^{i, syn} [z_{syn};\mathbbm{1}])
%     \tilde{K} &= (\sum_{i=1}^r W_{kv}^i [K;\mathbbm{1}]) \circ (\sum_{i=1}^r W_z^{i, sem} [z_{sem};\mathbbm{1}]) \\
%     \tilde{V} &= (\sum_{i=1}^r W_{kv}^i [V;\mathbbm{1}]) \circ (\sum_{i=1}^r W_z^{i, sem} [z_{sem};\mathbbm{1}])
% \end{aligned}
% % \right.
% \]
We refer \cite{liu-etal-2018-efficient-low} for an in-depth description of tensor fusion operations. 

\subsection{VAE Architecture} \label{sec:arc}

Finally, we integrate encoding and decoding phases into an end-to-end VAE architecture.

\paragraph{Encoder.} At the encoding stage, we consider the dual graph-sequential encoding mechanism adopting Bert as a sequential encoder and experimenting with two different graph-based encoders, including GraphSAGE \cite{hamilton2017inductive}, and Graph Transformer (TransCONV, \citet{yun2020graph}). To derive the syntactic space, $z_{syn}$, we use a mean pooling operation to obtain a sentence-level representation from the node embeddings $N$ and the adjacency matrix $A$: 
\[
\text{Embed}_{syn} = \text{MeanPool}(\text{GraphEnc}(A, N))
\]
For the semantic space, $z_{sem}$, we consider the special token [CLS] in BERT as the input of a linear transformation ($W$) to obtain a sentence-level representation: 
\[
\text{Embed}_{sem} = W(\text{LanguageEnc}(x)_{\text{[CLS]}})
\]
Where $x$ is the input sentence. Both spaces are constrained to follow a Gaussian distribution by learning the parameters $\mu$ and $\sigma$ through multilayer perceptions $W_\mu^{sem}$, $W_{\sigma}^{sem}$, $W_\mu^{syn}$, and $W_{\sigma}^{syn}$. The final latent representations can be obtained via:
\[
\begin{aligned}
    z_{sem(syn)} &= W_{\mu}^{sem(syn)} \times \text{Embed}_{sem(syn)} \\
    &+ W_{\sigma}^{sem(syn)}
\end{aligned}
\]
\paragraph{Decoder.} Because of the constraint of encoder architecture, the representations $z_{sem}$ and $z_{syn}$ have the potential to capture diverse features with a high level of disentanglement. To this end, we experiment with different decoding injection setups and low-rank operations (see Section \ref{sec:decoding_phase}) : (1) \emph{addition} for QKV (i.e., addition QKV), (2) \emph{fusion} for QKV (fusion QKV), (3) \emph{addition} for Q and \emph{memory} for KV (addition Q), and (4) \emph{fusion} for Q and \emph{memory} for KV (fusion Q).

\paragraph{Optimisation.} Our model can be trained via Evidence Lower Bound (ELBO) $x$ \cite{kingma2013auto}. To avoid the KL vanishing issue, which refers to the Kullback-Leibler (KL) divergence term in the ELBO becomes very small or approaches zero, we select the cyclical schedule to increase weights of KL $\beta$ from 0 to 1 \cite{fu-etal-2019-cyclical} and a KL thresholding scheme \cite{li-etal-2019-surprisingly} that chooses the maximum between KL and threshold $\lambda$. The final objective function can be described as follows:
\begin{align*} \label{eq:elbo_loss}
\mathcal{L}_\text{VAE} = & \mathbb{E}_{q^{sem}_\phi(z_{sem}|x), {q^{syn}_\phi(z_{syn}|A, N)}} \Big[ \log p_{\theta} \\
& ( x | z_{sem}, z_{syn}) \Big]  \\
& - \beta \max \left[ \lambda , \text{KL} q^{sem}_\phi(z_{sem}|x) || p(z) \right ] \\
& - \beta \max \left[ \lambda , \text{KL} q^{syn}_\phi(z_{syn}|x) || p(z) \right ]
\end{align*}
% Compared with previous work \cite{chen-etal-2019-multi} that learns the semantic-syntax information via multi-task loss terms and adversarial loss terms, we only use the ELBO loss function, which has the potential to deliver a stable training process.
% \input{tables_new/encoding_recon_cluster}
\paragraph{Information Bottleneck.} The dual graph-sequential encoding setup has the potential to alleviate information bottlenecks for sentence representations. In detail, \citet{li2020optimus} revealed that $\mathcal{L}_\text{VAE}$ is the upper bound of the information bottleneck (IB) (\textit{information bottleneck principle}, \citet{tishby2000information}).
\[
\mathcal{L}_\text{VAE} \ge (1-\beta) I_{q}(s,z) = \mathcal{L}_\text{IB}^{Bert}
\]
where $s$ and $z$ represent sentence and its corresponding latent representation $z$, $I_{q}$ is the mutual information, $q$ is encoder, $\mathcal{L}_\text{IB}$ is the Lagrange relaxation form \cite{tishby2000information}. As we mentioned in section \ref{sec:shared}, $s$ is composed of two kinds of information $\{x_{sem}\}$ and $\{x_{syn}\}$. In vanilla Optimus, $I(s,z)$ can be expanded into:
\[
\begin{aligned}
    I_q(s, z) &= I_{q}(x_{sem}+x_{syn};z)=I_{q}(x_{sem}, z) \\
    &+ I_{q}(x_{syn}, z)-I_{q}(x_{sem}, x_{syn}|z)
\end{aligned}
\]
Similarly, under the dual graph-sequential encoder setup, the mutual information can be described as:
\[
\mathcal{L}_\text{IB}^{Bert-graph} = I'_{q}(s, z) = I_{q}(x_{sem}, z) + I_{q}(x_{syn}, z)
\]
As we claimed before, $\{x_{sem}\} \cap \{x_{syn}\} \neq \emptyset$. Therefore, $\mathcal{L}_\text{IB}^{Bert} - \mathcal{L}_\text{IB}^{Bert-graph} = I_{q}(s,z)-I'_{q}(s,z) =- I_{q}(x_{sem}, x_{syn}|z) < 0$, indicating that the separated encoders can alleviate the information bottleneck. 

% The entropy of $s$ can be described as $H(s)=H(x_{sem} + x_{syn})$. With one single language encoder, the mutual information between $s$ and its latent representation $z$ can be described as:
% \[
% \begin{aligned}
% I(s; z) &= I(x_{sem}+x_{syn};z) \\
% &= H(x_{sem} + x_{syn}) - H(x_{sem} + x_{syn}|z)
% \end{aligned}
% \]
% Similarly, under graph-language encoders, the entropy of $s$ is $H(s)=H(x_{sem}) + H(x_{syn})$. The mutual information is:
% \[
% \begin{aligned}
% & I(x_{sem}; z) + I(x_{syn}; z) \\
% & = H(x_{sem}) + H(x_{syn}) - H(x_{sem}|z) - H(x_{syn}|z)
% \end{aligned}
% \]
% According to the \textit{subadditivity property} of entropy.
% \[
% \begin{aligned}
% H(x_{sem} + x_{syn}) &>= H(x_{sem}) + H(x_{syn}) \\
% H(x_{sem} + x_{syn}|z) &>= H(x_{sem}|z) + H(x_{syn}|z)
% \end{aligned}
% \]
% This is,
% \[
% \begin{aligned}
% I(s; z) 
% \end{aligned}
% \]
\section{Empirical Evaluation} \label{sec:empirical}

\begin{table*}[ht!]
% \scriptsize
% \setlength\tabcolsep{2.5pt}
\resizebox{15.6cm}{!}{
\small
\centering
\begin{tabular}{lcccc|cccccccc}
\toprule
Corpus & \multicolumn{4}{c|}{\textit{Mathematical expression}} & \multicolumn{5}{c}{\textit{Explanatory sentences}} \\
Proxy metrics & $\text{MSE}$(sem)$\downarrow$ & MSE(syn)$\downarrow$ & $\text{Acc}_{dep}$(syn)$\uparrow$ & $\text{Acc}_{dep}$(sem)$\downarrow$  & $\text{MSE}$(sem)$\downarrow$ & MSE(syn)$\downarrow$ & $\text{Acc}_{dep}$(syn)$\uparrow$ & $\text{Acc}_{dep}$(sem)$\downarrow$ & $\text{F1}_{dep}$(sem)$\downarrow$\\ \midrule 
LSTM & 079.02 & 070.48 & 000.74 & 000.74 & 176.39 & 158.03 & 000.40 & 000.40 & 000.41\\
VGAE & 125.68 & 434.52 & 000.81 & 000.82 & 169.42 & 110.30 & 000.40 & 000.38 & 000.45 \\
Siam & 191.97 & 053.90 & 000.85 & 000.52 & 074.86 & 031.95 & 000.43 & 000.35 & 000.42 \\
GraphEncoder & -- & -- & -- & -- & -- & -- & -- & -- & -- \\
 + GCN & \underline{\textbf{\textcolor{blue}{004.31}}} & 065.79 & 000.72 & \underline{\textbf{\textcolor{blue}{000.27}}} & 069.77 & 091.94 & 000.49 & \underline{\textbf{\textcolor{blue}{000.12}}} & \underline{\textbf{\textcolor{blue}{000.30}}} \\
 + GraphSAGE & 208.21 & 053.20 & \underline{\textbf{\textcolor{blue}{000.98}}} & 000.52 & 058.12 & 004.10 & 000.50 & 000.39 & 000.46  \\
 + TransConv & 249.00 & \underline{\textbf{\textcolor{blue}{038.30}}} & \underline{\textbf{\textcolor{blue}{000.98}}} & 000.57 & \underline{\textbf{\textcolor{blue}{058.10}}} & \underline{\textbf{\textcolor{blue}{003.35}}} & \underline{\textbf{\textcolor{blue}{000.51}}} & 000.38 & 000.47 \\ \midrule
$\text{F1}_{dep}^*$(sem)$\downarrow$ & $\text{F1}_{dep}$(syn)$\uparrow$ & MI(sem,syn)$\downarrow$ & KL(sem||syn)$\uparrow$ & Wass(sem,syn)$\uparrow$ & $\text{F1}_{dep}$(syn)$\uparrow$ & MI(sem,syn)$\downarrow$ & KL(sem||syn)$\uparrow$ & Wass(sem,syn)$\uparrow$\\ \midrule
\multicolumn{1}{c}{000.71} & 000.70 & 004.88 & 005.74 & 000.53 & 000.43 & 004.87 & 001.01 & 000.78\\
\multicolumn{1}{c}{000.84} & 000.84 & 004.85 & 026.12 & 000.32 & 000.44 & 004.66 & 007.04 & 000.90 \\
\multicolumn{1}{c}{000.41} & 000.87 & 004.85 & 011.95 & 000.69 & 000.44 & 004.96 & 008.72 & 000.80\\
\multicolumn{1}{c}{--} & -- & -- & -- & -- & -- & -- & -- & --\\
\multicolumn{1}{c}{\underline{\textbf{\textcolor{blue}{000.24}}}} & 000.79 & 004.82 & 024.05 & 000.72 & \underline{\textbf{\textcolor{blue}{000.54}}} & 004.78 &  011.77 & 000.30 \\
\multicolumn{1}{c}{000.42} & \underline{\textbf{\textcolor{blue}{000.98}}} & 005.04 & 005.12 & 000.69 & 000.44 & 004.45 & \underline{\textbf{\textcolor{blue}{043.45}}} & \underline{\textcolor{blue}{\textbf{001.92}}} \\
\multicolumn{1}{c}{000.52} & \underline{\textbf{\textcolor{blue}{000.98}}} & \underline{\textbf{\textcolor{blue}{004.80}}} & \underline{\textbf{\textcolor{blue}{031.63}}} & \underline{\textbf{\textcolor{blue}{001.19}}} & 000.48 & \underline{\textbf{\textcolor{blue}{003.54}}} & 012.78 & 000.75\\ \bottomrule
\end{tabular}
}
\caption{Proxy metrics for evaluating the organisation of the latent syntactic and semantic space for different encoding configurations of Optimus. The best \underline{\textbf{\textcolor{blue}{results}}} indicate that graph-language encoding setup can efficiently capture syntax information and maintain separation.} \label{tab:enoding_syntax}
\end{table*}

Following the stages in our methodology, we first evaluate different encoding setups for injecting syntactic information into VAEs (as illustrated in Section \ref{sec:encoding_phase}). Subsequently, we consider the best encoding configuration to examine which decoding strategy (as illustrated in Section \ref{sec:arc}) can lead to better language modelling performances. Finally, we evaluate the best architectural setup for downstream tasks.

To experiment, we focus on both natural and formal languages, training the models on \emph{explanatory sentences} and \emph{mathematical expressions}. The rationale behind this choice is that (1) explanatory sentences \cite{jansen2018worldtree,valentino2022hybrid,thayaparan-etal-2021-explainable, zhang2023type} provide a semantically challenging yet sufficiently well-scoped scenario to evaluate the syntactic and semantic organisation of the space; (2) mathematical expressions \cite{valentino2023multioperational,meadows2023symbolic} follow a well-defined syntactic structure and set of symbolic rules that are notoriously difficult for neural models. Moreover, the set of rules applicable to a mathematical expression fully determines its semantics, allowing for an in-depth inspection and analysis of the precision and level of generalisation achieved by the models. All experimental details are provided in Appendix \ref{sec:enc_baselines}.

% In this paper, we focus on Math expression \cite{meadows2023symbolic} since it has regular semantic and syntax features. As for baselines, instead of the aforementioned setups in section \ref{sec:related}, we also select vanilla Optimus \cite{li2020optimus} and four LSTM-based autoencoders (AEs), including $\beta$-VAE \cite{Higgins2016betaVAELB}, adversarial AE (\citet{makhzani2016adversarial}, AAE), label adversarial AE (\citet{rubenstein2018latent}, LAAE), and denoising adversarial autoencoder (\citet{shen2020educating}, DAAE). All baselines have a latent size of 768. For semantic-syntax separated VAE setups, we evenly split the latent space for both.

% As for natural language corpus, we quantitatively evaluate the performance of the models using five metrics, including BLEU \cite{Papineni02bleu:a}, BLEURT \cite{https://doi.org/10.48550/arxiv.2004.04696}, cosine similarity from pre-trained sentence T5 \cite{https://doi.org/10.48550/arxiv.2108.08877}, cross-entropy (Loss), and perplexity (PPL). For the mathematical expression corpus, we evaluate the robustness and generalization of models on out-of-distribution (OOD) test sets.

\subsection{Encoding: Latent Representations} \label{sec:enc_syn}
\paragraph{Evaluation.} Firstly, we evaluate different encoding setups to the effect of semantic-syntax distribution in latent space from three perspectives: (i) latent space geometry: whether the latent space can capture the corresponding features -- i.e., sentences with the same/different features are clustered/separated accordingly in the latent space. In this case, we can evaluate the organisation of the latent space via MSE of k-mean \cite{zhang2022quasi, zhang2023learning, michlo2023overlooked}, (ii) syntactic features: following the probing method \cite{conneau-etal-2018-cram}, we train a linear classifier to predict tree depth. Here, better classification performances indicate a higher separability of syntactic features in the latent space, and (iii) semantic and syntax space alignment: we adopt statistical metrics to compare latent distributions such as Mutual Information (MI), Kullback–Leibler divergence (KL), and Wasserstein distance (Wass). As illustrated in Table \ref{tab:enoding_syntax}, we can observe that (1) the Optimus(GraphEncoder) can better capture the syntactic structures and induce a better latent space separation, (2) It can lead to a better organisation of the semantic space $\text{MSE}(sem)$. We will further explore this phenomenon in subsequent sections.

\begin{figure}[ht!]
    \centering
    \includegraphics[width=\columnwidth]{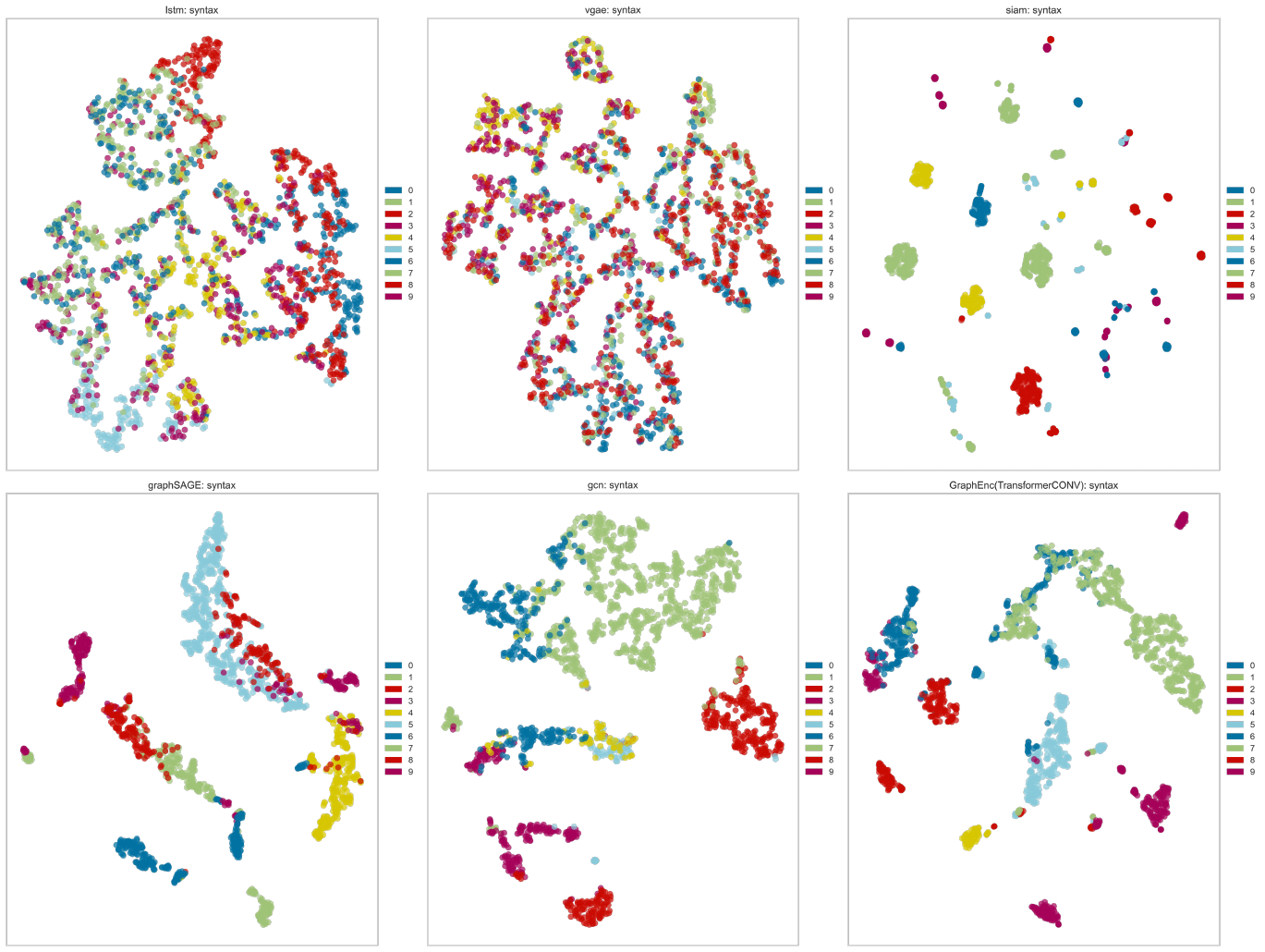}
    \caption{Visualizing the syntactic clusters for mathematical expressions reveals that graph encoder can better represent syntax information in latent sentence space (top: LSTM, VGAE, Siam, bottom: graph encoders with GraphSAGE, GCN, TransformerCONV).}
    \label{fig:syntax_latent_space}
\end{figure}
\begin{table*}[ht!]
\scriptsize
\setlength\tabcolsep{2.5pt}
\resizebox{15.6cm}{!}{
\small
\centering
\renewcommand\arraystretch{1}
\begin{tabular}{lccccccccccccccc}
\toprule
\multicolumn{1}{c}{Corpus} & \multicolumn{10}{c}{\textit{Mathematical expression}} & \multicolumn{5}{c}{\textit{Explanatory sentences}} \\ 
\multicolumn{1}{c}{Metrics} & \multicolumn{2}{c}{EVAL} & \multicolumn{2}{c}{VAR-SWAP} & \multicolumn{2}{c}{EASY} & \multicolumn{2}{c}{EQ-CONV} & \multicolumn{2}{c}{LEN} & BLEU & BLEURT & Cosine & Loss$\downarrow$ & PPL$\downarrow$ \\ \midrule
% Metrics & EM & Bleu & EM & Bleu & EM & Bleu & EM & Bleu & EM & Bleu \\ \hline

\multicolumn{16}{c}{\textit{sentence VAE baselines}} \\
01. AAE(768) & 0.10 & 0.75 & 0.00 & 0.25 & 0.02 & 0.53 & 0.00 & 0.54 & 0.00 & 0.51 & 0.35 & -0.95 & 0.80 & 3.35 & 28.50\\
02. LAAE(768) & 0.00 & 0.43 & 0.00& 0.25 & 0.00 & 0.27 & 0.00 & 0.29 & 0.00 & 0.44 & 0.26 & -1.07& 0.78& 3.71 & 40.85\\
03. DAAE(768) & 0.00 & 0.24 & 0.00& 0.21 & 0.00 & 0.21 & 0.00 & 0.22 & 0.00 & 0.42 & 0.22 & -1.26& 0.76& 4.00 & 54.59\\
04. $\beta$-VAE(768) & 0.00 & 0.14 & 0.00& 0.15 & 0.00 & 0.13 & 0.00 & 0.14 & 0.00 & 0.35 & 0.06& -1.14& 0.77& 3.69 & 40.04\\ 
05. Optimus(768) & 0.99 & 0.99 & 0.00& \underline{\textbf{\textcolor{blue}{0.38}}} & 0.81 & 0.93 & 0.00& 0.81 & \underline{\textbf{\textcolor{blue}{0.14}}} & 0.76 & 0.35 & -0.59 & 0.83 & 0.98 & 2.66 \\ \midrule %& 0.42 & -0.24 & 0.87 & 0.63 & 1.87\\ \hline \hline % \hdashline
% \multicolumn{16}{c}{\textit{Semantic-syntax disentanglement baselines}} \\
% DSS-VAE & \\
% VGVAE & \\ \hline \hline
\multicolumn{16}{c}{\textit{different encoding setups with memory injection}} \\
06. LSTM & \underline{\textbf{\textcolor{blue}{1.00}}} & \underline{\textbf{\textcolor{blue}{1.00}}} & 0.00& 0.35 & 0.73 & 0.94 & 0.00& 0.77 & 0.06 & 0.74 & 0.41 & -0.41 & 0.85 & 1.04 & 2.82 \\ % 0.45 & -0.17 & 0.88 & 0.64 & 1.89 \\
07. VGAE & 0.98 & 0.99 & 0.00& 0.34 & 0.72 & 0.93 & 0.00& 0.74 & 0.04 & 0.71 & 0.26 & -0.91 & 0.78 & 1.14 & 2.55 \\
08. Siam & \underline{\textbf{\textcolor{blue}{1.00}}} & \underline{\textbf{\textcolor{blue}{1.00}}} & 0.00& 0.30 & 0.22 & 0.80 & 0.00& 0.78 & 0.03 & 0.75 & 0.49 & -0.15 & 0.88 & 0.94 & 2.55 \\ % \textbf{\textcolor{blue}{0.57}} & \textbf{\textcolor{blue}{0.12}} & \textbf{\textcolor{blue}{0.91}} & \textbf{\textcolor{blue}{0.68}} & \textbf{\textcolor{blue}{1.97}} \\ % 0.57 & 0.15 & 0.91 & \\ %0.57 & 1.76 \\
GraphEncoder \\ 
09. + GCN & 0.00 & 0.40 & 0.00& 0.22 & 0.00 & 0.27 & 0.00& 0.37 & 0.00 & 0.43 & 0.15 & -1.19 & 0.75 & 1.24 & 3.45 \\
10. + GraphSAGE & 0.88 & 0.96 & 0.00& 0.28 & 0.06 & 0.46 & 0.00& 0.69 & 0.00 & 0.60 & 0.45 & -0.28 & 0.87 & 1.00 & 2.71 \\
11. + TransCONV & 0.89 & 0.95 & 0.00& 0.28 & 0.14 & 0.53 & 0.00 & 0.67 & 0.00 & 0.61 & 0.17 & -1.16 & 0.75 & 1.21 & 3.35 \\ \midrule
\multicolumn{16}{c}{\textit{Graph-language encoders: injecting syntax into Q, semantic into KV}} \\
Bert-GraphSAGE & \\
% + fuse sem-syn & \\
% + sep dec & \\
12. + addition Q & 0.99& 0.99 & 0.00 & 0.27 & 0.23& 0.63& 0.00& 0.71 &0.02&0.66& 0.60 &0.22&0.92 &0.74 & 2.09 \\
13. + addition QKV & \underline{\textbf{\textcolor{blue}{1.00}}} & \underline{\textbf{\textcolor{blue}{1.00}}} & 0.00 & 0.35 & 0.65 & 0.90 &0.00& 0.80 & 0.06 & 0.75 & 0.63 & 0.31 & 0.93 & 0.65 & 1.91 \\
14. + fusion Q & 0.94 & 0.97 & 0.00 & 0.29 & 0.08 & 0.63 & 0.00 & 0.71 & 0.00 & 0.62 & 0.55 & 0.03 & 0.91 & 0.90 & 2.45 \\ % 0.52 & -0.01 & 0.90 & 1.11 & 3.03 \\
15. + fusion QKV & \underline{\textbf{\textcolor{blue}{1.00}}} & \underline{\textbf{\textcolor{blue}{1.00}}} &0.00& \underline{\textbf{\textcolor{blue}{0.38}}} & 0.37 & 0.84 & 0.00 & 0.80 & 0.02 & 0.73 & 0.46 & -0.23 & 0.88 & 1.10 & 3.00 \\
Bert-TransCONV & \\
% + fuse sem-syn & 0.00 & 0.50 & 0.00 & 0.20 & 0.00 & 0.37 & 0.00 & 0.44 & 0.00 & 0.44 \\
% + sep dec & \\
16. + addition Q & 0.98& 0.99 &0.00& 0.26&0.31&0.69&0.00&0.67&0.01&0.63& 0.59 & 0.18 & 0.92 & 0.76 & 2.13 \\
17. + addition QKV & \underline{\textbf{\textcolor{blue}{1.00}}} & \underline{\textbf{\textcolor{blue}{1.00}}} & 0.00 & \underline{\textbf{\textcolor{blue}{0.38}}} & \underline{\textbf{\textcolor{blue}{0.90}}} & \underline{\textbf{\textcolor{blue}{0.98}}} &0.00 & \underline{\textbf{\textcolor{blue}{0.82}}} & 0.10 & \underline{\textbf{\textcolor{blue}{0.78}}} & \underline{\textbf{\textcolor{blue}{0.65}}} & \underline{\textbf{\textcolor{blue}{0.35}}} & \underline{\textbf{\textcolor{blue}{0.94}}} & \underline{\textbf{\textcolor{blue}{0.62}}} & \underline{\textbf{\textcolor{blue}{1.85}}} \\
18. + fusion Q & 0.96 & 0.98 & 0.00 & 0.29 & 0.18 & 0.60 & 0.00 & 0.74 & 0.00 & 0.64 & 0.53 & -0.02 & 0.90 & 0.98 & 2.66 \\ 
19. + fusion QKV & 0.99 & 0.99 & 0.00 & 0.35 &0.45 & 0.82 & 0.00 & 0.80 & 0.01 & 0.74 & 0.46 & -0.16 & 0.88 & 1.13 & 3.09\\ \bottomrule
\end{tabular}
}
\caption{Results on language modelling. Regarding mathematical expressions, we adopt exact match (left) and bleu (right) as evaluation metrics for each test set. The best results are highlighted in \underline{\textbf{\textcolor{blue}{blue}}}.} \label{tab:enoding_recon}
\end{table*}
\paragraph{Visualisation.} Next, we visualize the cluster separation of latent space via t-SNE \cite{JMLR:v9:vandermaaten08a} (see Figure \ref{fig:syntax_latent_space}). From the visualisation, we can observe that the Optimus(GraphEncoder) can induce a better separation between different syntactic clusters. We also provide a qualitative evaluation by decoding the latent representation of each cluster and visualisation for explanatory sentences in Appendix \ref{sec:exp_res}. These results allow us to conclude that the integration of graph-based and sequential models in a dual encoder configuration can better capture structural syntactic information while maintaining a separation between latent spaces.

\subsection{Decoding: Language Modelling} \label{sec:dec_lm}

\paragraph{Baselines.} We assess performances on language modelling using a different set of baselines. Specifically, we evaluate the performance of vanilla Optimus \cite{li2020optimus} and four LSTM-based autoencoders (AEs), including $\beta$-VAE \cite{Higgins2016betaVAELB}, adversarial AE (\citet{makhzani2016adversarial}, AAE), label adversarial AE (\citet{rubenstein2018latent}, LAAE), and denoising adversarial autoencoder (\citet{shen2020educating}, DAAE). All baselines have a latent size of 768. For semantic-syntax separated VAE setups, we evenly split the latent space for both. Moreover, we compare the proposed injection mechanism via low-rank operations with a standard memory injection setup \cite{li2020optimus}.

\paragraph{Metrics.} As for modelling mathematical latex expressions, we use Exact Match (EM) and Bleu to evaluate the robustness of models on five different test sets where four of them include out-of-distribution examples, including (1) EVAL: mathematical expressions following the same distribution as the training set (like $U + cos{(n)}$), (2) VAR: mathematical expressions composed of a different set of variables (like $U + cos{(beta)}$), (3) EASY: simpler mathematical expressions with a lower number of variables (like $cos{(n)}$), (4) EQ: mathematical derivations with equality insertions (like $E = U + cos{(n)}$), (5) LEN:  mathematical derivations with a higher number of variables (like $U + cos{(n)}) + A + B$). Regarding explanatory sentences, we evaluate the performance of the models using five metrics, including BLEU \cite{Papineni02bleu:a}, BLEURT \cite{https://doi.org/10.48550/arxiv.2004.04696}, cosine similarity from pre-trained sentence T5 \cite{https://doi.org/10.48550/arxiv.2108.08877}, cross-entropy (Loss), and perplexity (PPL).

\paragraph{Results.} Firstly, we evaluate the performance of baselines with different syntactic injection setups. As illustrated in the middle part of Table \ref{tab:enoding_recon}, we can find that most configurations lead to lower performance, especially when using graph encoders, compared to vanilla Optimus, indicating that a standard memory injection mechanism for leveraging heterogeneous latent space is not effective. Conversely, by comparing line 05 to lines 12, 14, 16 and 18, it is possible to notice that injecting only syntactic information in Q can improve reconstruction performances on explanatory sentences. Moreover, we evaluate whether injecting heterogeneous latent representations into different attention components (Q, K, V) can further improve the results. In the bottom part of Table \ref{tab:enoding_recon}, we find that injecting semantic and syntax spaces into different attention components can additionally improve model performance (lines 9-11 vs 12, 14, 16, 18), demonstrating that semantic and syntax space possess complementary features. Finally, we evaluate which injection strategies can achieve the best results. We found that \textit{addition} injection with Bert-TransCONV (line 17) can achieve the best overall results on both corpora. Next, we perform a further analysis attempting to explain why syntax injection can improve model performance, especially on natural language sentences. 
\begin{figure}[ht!]
    \centering
    \includegraphics[width=\columnwidth]{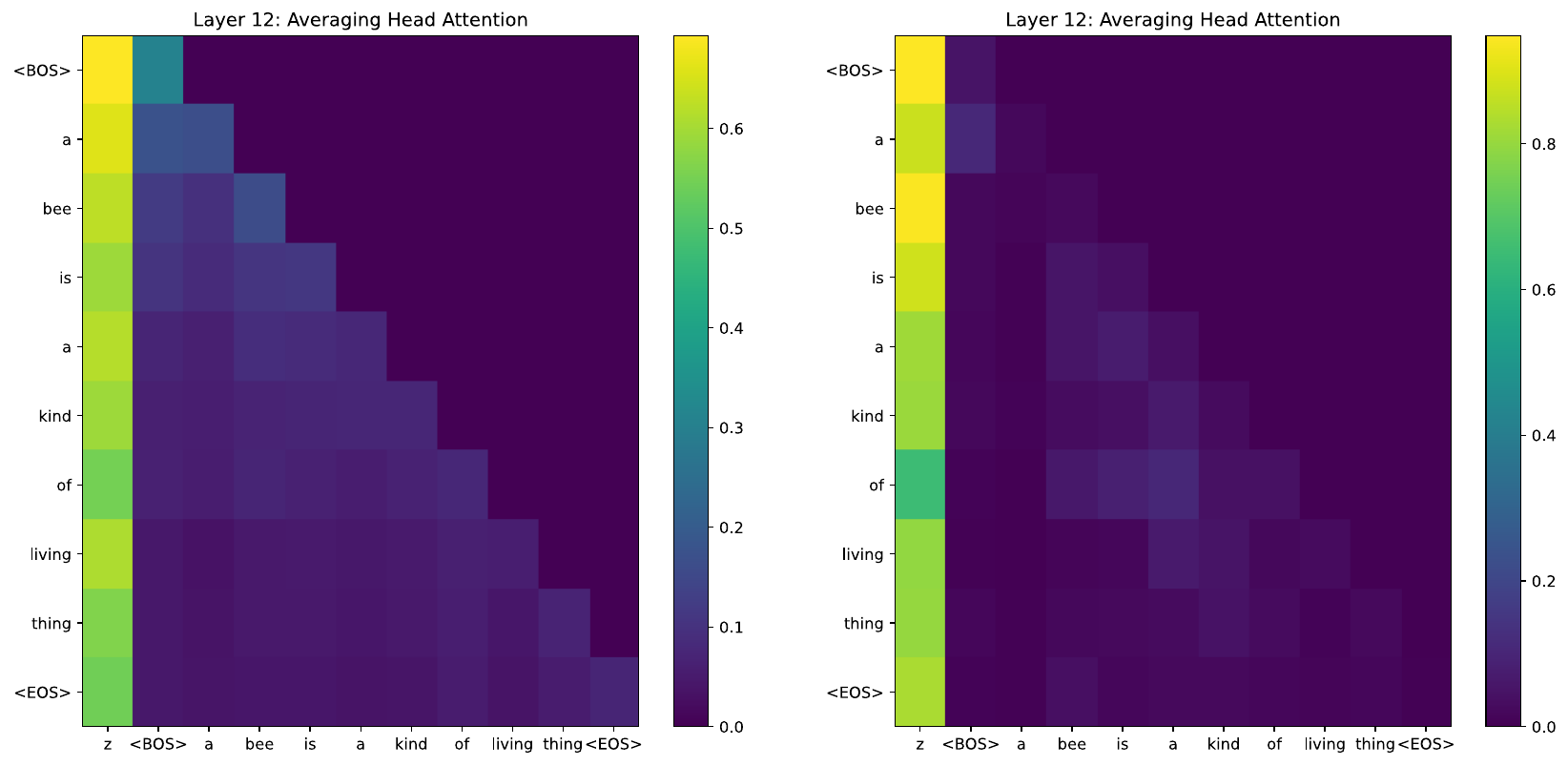}
    \caption{Visualizing attention weighs (left: vanilla Optimus, right: Bert-TransCONV with \textit{addition Q} setup) where \textit{bee}: 0.58 < 0.94, \textit{living thing}: (0.27, 0.15) < (0.80, 0.80).}
    \label{fig:attn}
\end{figure}
\paragraph{Analysis.} Under the VAE architecture, we conjecture that the latent space separation between syntax and semantics allows the BERT encoder to capture and represent more fine-grained and accurate word-level information, alleviating the loss of information in the sentence bottleneck. To support this hypothesis, We provide a set of qualitative examples in Table \ref{tab:rec}. Here, we report some representative examples. Given the target input: \textit{a bee is a kind of living thing}, we found the reconstruction of vanilla Optimus to be \textit{a frog is a kind of amphibian}. This shows that Optimus is distracted by syntactic features, (\textit{x is a kind of y}) that are highly frequent in the training set, and struggles in the reconstruction of specific lexical content (i.e., \textit{frog} and \textit{amphibian}). In contrast, we found that the proposed architecture allows the semantic space to specialise in lexical content since the graph-based encoder can already capture fine-grained syntactic information. To additionally support such a claim, we investigate the heatmap of the attention weights of GPT2. In figure \ref{fig:attn}, the first column of each heatmap represents the lexical information carried by the latent representation. Here, we can observe that the proposed architecture with Bert-TransCONV + \textit{addition Q} setup (right) can pay more attention to specific lexical elements (i.e., \textit{bee}) compared to vanilla Optimus (left). This also explains how the integration of a graph-based encoder can indirectly lead to improved organisation for the semantic space (see MSE in Table \ref{tab:enoding_syntax}). We provide additional examples of attention heatmaps in Appendix \ref{sec:attn_map}.

% From the perspective of information theory, 
% \textcolor{red}{(can we explain from the perspective of information theory, information bottleneck?)}

% \textit{2. why did the ``addition'' injection leads to better reconstruction?} As for \textit{addition} semantic injection, each value with index $i,j$ in $\tilde{Q} \times \tilde{K}^T$ weight matrix can be described as:
% \[
% \begin{aligned}
%     v_{ij} &= (q_{i1}+z^{syn}_1) \times (k_{j1} + z^{sem}_1) + \dots \\
%     &+ (q_{in}+z^{syn}_n) \times (k_{jn} + z^{sem}_n)
% \end{aligned}
% \]
% Where $z^{syn(sem)} = [z^{syn(sem)}_1, \dots, z^{syn(sem)}_n]$, $q_{in}$ and $k_{jn}$ are the weights in Q and K matrix, $n$ is 64 in GPT2 attention. As we can observe, the semantic information is injected into each value of the attention weight matrix. In \textit{memory} semantic injection, however, the $z^{sem}$ is only injected into the first column of $\tilde{Q} \times \tilde{K}^T$ weight matrix, that is,
% \[
% \begin{aligned}
%     v_{i1} &= (q_{i1}+z^{syn}_1) \times z^{sem}_1 + \dots \\
%     &+ (q_{in}+z^{syn}_n) \times z^{sem}_n
% \end{aligned}
% \]
% which are weakly connected to the $z^{syn}$ and $K$ and are likely to be ignored by the decoder, exacerbating the KL vanishing.

\subsection{Downstream Evaluation}
\begin{figure}[t]
    \centering
    \includegraphics[width=\columnwidth]{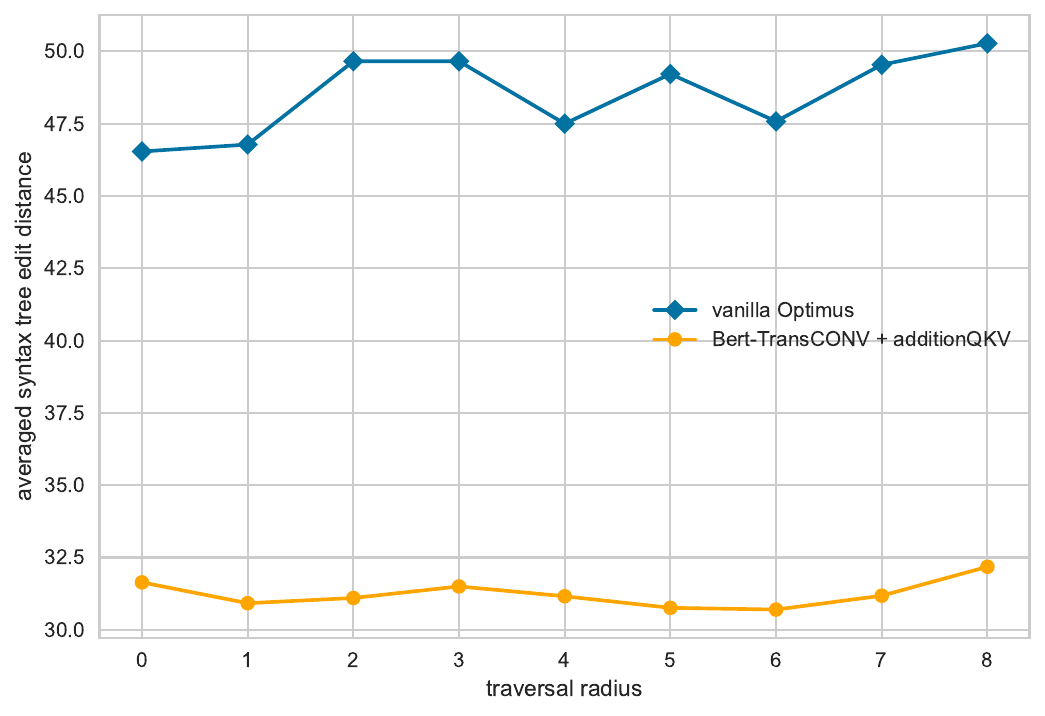}
    \caption{Traversing semantic space with increasing traversal radius while keeping syntax space fixed. We can observe improved syntax control in decoding by separating syntax and semantics.}
    \label{fig:trav}
\end{figure}

\paragraph{Guided Generation.} One advantage of the VAE architecture is that it allows controlling sentence generation by manipulating latent representations via traversal, interpolation, and vector arithmetic \cite{shen2020educating,zhang2023type}. By leveraging such property, we quantitatively assess the controllability of the decoding via latent traversal. Specifically, given an input sentence as an initial point, we perform an \textit{Ornstein-Uhlenbeck} random walk \cite{pinsky2010introduction} \footnote{$\tilde{z}_{t+1}=-\gamma \tilde{z}_{t}+\sigma W_t$ where $t$ is the index, $W_t \in N(0,1)$, $\gamma$ and $\sigma$ are scalar hyper-parameters.} for semantic space and fix syntax space. If the model can disentangle semantic and syntactic features, we expect the generated sentence to change lexical content while keeping a fixed syntactic structure. To experiment, we quantitatively evaluate the similarity of syntactic structures between input and traversed sentences via syntax tree edit distance. We gradually increase the radius of the random walk to perform a comparison between vanilla Optimus and Bert-TransCONV(addition QKV) (see Figure \ref{fig:trav}). From the results, we can conclude that the proposed architecture can hold the syntax structure unchanged, indicating better controllability and separation. We provide qualitative examples of such behaviour in Appendix \ref{sec:app_traversal}.

% Besides, we also provide the traversed sentences in table \ref{tab:traversal_example}. From it, we can observe that the semantic-syntax separation can better hold the syntax structures of the traversed sentences and have the potential to hold similar semantics, indicating the graph-induced latent semantic geometry is more regular than that of normal Optimus, resulting in better controllability during decoding. We also provide the traversed examples of syntax space in table \ref{tab:traversal_example_1}. From it, we can observe that the generated sentences hold similar semantics about \textit{sea} and \textit{water} as input, compared to normal Optimus which generates unrelated semantics: \textit{desert} and \textit{forest}, etc.
% \input{tables_new/traversal_example}
% \input{tables_new/traversal_example_1}

\paragraph{Mathematical Derivations.} Finally, we explore the quality of the representation for mathematical expressions on downstream equational inference tasks \cite{meadows2023symbolic,valentino2023multioperational}. Specifically, we focus on expression derivation, where, given a premise $x$ and a mathematical operation $t$ (i.e., differentiation, integration) the goal is to predict whether a target mathematical expression $y$ can be derived from $x$ via $t$. Here, we adopt the dataset introduced by \citet{valentino2023multioperational} and examine whether a linear probing classifier \cite{ferreira2021does} trained on latent expression representations encoded from frozen pre-trained models (i.e., concatenating syntactic and semantic space), can predict the correct operation $t$ (i.e., Operation Classification) in a multi-label classification problem (i.e., given premise $x$ and target result $y$) and whether the classifier can predict a valid conclusion $y$  (i.e. Conclusion Classification) given a premise $x$ in a binary classification setting (using random negative examples). Experimental results reveal that separately injecting latent semantic and syntactic representations can provide complementary information and improve performance on both probing tasks.

\begin{table}[t]
\scriptsize
\setlength\tabcolsep{2.5pt}
\resizebox{7.8cm}{!}{
\small
\centering
\renewcommand\arraystretch{1}
\begin{tabular}{lcccc}
\toprule
Inference Type & \multicolumn{2}{c}{Operation Class.} & \multicolumn{2}{c}{Conclusion Class.} \\ 
Metrics & Acc & F1  & Acc & F1\\ \midrule
Optimus(768) & 0.89 & 0.89 & 0.68 & 0.68 \\
\midrule
LSTM & 0.89 & 0.89 & 0.59 & 0.62 \\
VGAE & 0.79 & 0.80 & 0.56 & 0.62 \\
Siam & \underline{\textbf{\textcolor{blue}{0.92}}} & \underline{\textbf{\textcolor{blue}{0.92}}} & 0.59 & 0.59 \\
GraphEncoder &  &  &  &  \\ 
+ GCN & 0.73 & 0.74 & 0.57 & 0.55  \\
+ GraphSAGE & 0.87 & 0.87 & 0.64 & 0.63 \\
+ TransCONV & 0.88 & 0.89 & 0.63 & 0.62 \\ \midrule
Bert-GraphSAGE & \\
+ addition QKV & 0.88 & 0.88 & 0.69 & 0.69 \\
+ fusion QKV & 0.90 & 0.90 & \underline{\textbf{\textcolor{blue}{0.71}}} & \underline{\textbf{\textcolor{blue}{0.71}}}  \\
Bert-TransCONV & \\
+ addition QKV & \underline{\textbf{\textcolor{blue}{0.92}}} & \underline{\textbf{\textcolor{blue}{0.92}}} & 0.68 & 0.68 \\ 
+ fusion QKV & 0.91 & 0.91 & 0.59 & 0.59 \\ \bottomrule
\end{tabular}
}
\caption{Results for the mathematical derivations probing task reveal that separately injecting latent semantic and syntactic representations can provide complementary information, resulting in enhanced performance.} \label{tab:classification}
\end{table}

\section{Related work} \label{sec:related}
\paragraph{Language VAE.} Most previous language VAE works are based on LSTM architectures instantiated on different text generation tasks, including story generation \cite{fang2021transformerbased}, dialogue generation \cite{zhao-etal-2017-learning}, text style transfer \cite{john-etal-2019-disentangled, shen2020educating}, text paraphrasing \cite{bao-etal-2019-generating}, among others. The development of Optimus \cite{li2020optimus} led to an increasing amount of research focusing on how to control the generation of Transformer-based architectures (i.e., Bert-GPT2 setup) by latent space geometry \cite{zhang2022quasi, zhang2023learning} or pre-defined priors \cite{fang-etal-2022-controlled}. Comparatively, this work contributes to improving the semantic-syntax separation with the help of a graph-based encoder. Combining a Graph Encoder with a sequential language decoder has been deployed in different domains, such as Abstract meaning representation (AMR)-to-Text, \citet{wang-etal-2020-amr}, knowledge graphs (KG)-to-Text \cite{koncel-kedziorski-etal-2019-text,ribeiro-etal-2021-investigating}. However, to our knowledge, the combination of a graph encoder and a language VAE architecture for text generation is still underexplored.

\paragraph{Learning Syntactic Representations.} From the perspective of model architecture, three kinds of encoders can be used to learn syntactic representations, including graph-based encoders \cite{wu2023graph}, sequential encoders \cite{vaswani2017attention}, and tree-based encoders \cite{harer2019tree}, with the latter two commonly used in the natural language generation domain \cite{Raffel2020t5}. Sequential encoders, such as LSTMs \cite{10.1162/neco.1997.9.8.1735} and Transformers \cite{vaswani2017attention}, have demonstrated to capture syntax features when trained on flattened syntax trees \cite{bao-etal-2019-generating, https://doi.org/10.48550/arxiv.2104.08661}. On the other side, tree-based encoders can capture the structural representations by recursively modelling each node in a tree -- i.e., using Recursive Neural Networks \cite{harer2019tree, mrini-etal-2021-recursive}. Nevertheless, whether these models truly capture structural information or just the lexical combination of tokens is not fully clarified \cite{shi2016does}. In this work, we propose the use of graph-based encoders, such as Graph Convolutional Networks (GCN) \cite{kipf2016semi}, to better capture structural information and topological relations in syntactic trees. Graph Neural Networks (GNNs) \cite{zhou2020graph} have been effective for encoding explicit syntactic and relational structures in various NLP tasks \cite{wu2023graph,sachan-etal-2021-syntax,veyseh2020graph,theodoropoulos-etal-2021-imposing}.

%between nodes can evidently present the structural representation. Therefore, this work selects the graph-based encoders as the syntax encoders.

\section{Conclusion} \label{sec:concl}   
In this work, we focused on the semantic-syntax separation through language VAEs, especially Optimus (Bert-GPT2), architecture. We first implement several encoding baselines and reveal that language-graph encoding setups can better capture syntax information and maintain semantic-syntax separation. However, the language-graph encoding setup leads to low reconstruction performance. To solve this problem, we explored the integration of heterogeneous latent spaces via injection mechanisms. Experimental results showed that our setup can greatly improve language modelling performance, and revealed that the semantic-syntax separation can assist the language modelling task since the language encoder pays more attention to fine-grained lexical semantics, avoiding the distraction of syntax information captured by the separated syntax encoder, which can alleviate the information bottleneck of the language encoder.
\section{Limitations} \label{sec:limit}
Although the semantic-syntax separated latent space can provide better latent space geometry, how can we efficiently control the decoding stage through latent geometry is still challenging due to the discrete nature of the latent sentence space. Besides, robustness towards out-of-distribuion generalization for within the separated latent spaces will be further investigated.

\section*{Acknowledgements}
This work was partially funded by the Swiss National Science Foundation (SNSF) project NeuMath (200021\_204617) and by the Engineering and Physical Sciences Research Council (EPSRC) under Grant EP/T026995/1 EnnCore.

\bibliography{references}
\bibliographystyle{acl_natbib}

\appendix
\clearpage
% \appendix

\section{Training setups} \label{sec:enc_baselines}
\paragraph{Datasets} Table \ref{tab:stats_data} displays the statistical information of the datasets used in the experiment. As for the AutoEncoder setup, we use the non-repetitive explanations selected from both WorldTree \cite{jansen2018worldtree} and EntailmentBank \cite{https://doi.org/10.48550/arxiv.2104.08661} corpus as the experimental data. The mathematical expressions are derived from \cite{meadows2023symbolic}.
\begin{table}[ht!]
    \small
    \centering
    \renewcommand\arraystretch{1}
      % \resizebox{7.6cm}{!}{
    \begin{tabular}{|c|cc|}
        \hline
        Corpus & Num data. & Avg. length \\ \hline
        WorldTree & 11430 & 8.65 \\
        EntailmentBank & 5134 & 10.35 \\ 
        Math Symbol & 32000 & 6.84 \\ \hline
        
    \end{tabular}
    % }
    \caption{Statistics from datasets.} \label{tab:stats_data}
\end{table}
\paragraph{Tokenization} As for mathematical expression, we add specific math tokens, including $\text{frac}, \sin, \cos, \log, e$, into the dictionary of both Bert and GPT2 and consider the remaining tokens as char-level. As for explanatory sentences, we use the default tokenization in Bert and GPT2.

\paragraph{Syntax parsing} As for mathematical expression, we use Expression Trees \footnote{\url{https://docs.sympy.org/latest/tutorials/intro-tutorial/manipulation.html}}, As for explanatory sentences, we use consistency parser\footnote{\url{https://demo.allennlp.org/constituency-parsing}} from AllenNLP library \cite{gardner2018allennlp} to get the flattened syntax tree, and remove all word content from the tree as the input of graph encoder.
\paragraph{Model implementation} As for graph encoders, we use \textit{PyTorch Geometric} library \footnote{\url{https://pytorch-geometric.readthedocs.io/en/latest/}}. We deployed two hidden layers for GCN, GraphSAGE, and TransformerCONV. For mathematical expression, we replace the content of variables with random noises following uniform distribution with the range between -1 and 1 during the node embedding stage. The implementation of Optimus is based on their original code \footnote{\url{https://github.com/ChunyuanLI/Optimus}}. The implementation of LSTM-based VAEs is based on the code supplied from \citet{shen2020educating} \footnote{\url{https://github.com/shentianxiao/text-autoencoders}}.

\paragraph{Hyperparameters} In the experiment, all baselines and our architecture hold the same size of latent representation (768). The training epoch is 100, the learning rate is 5e-5, the batch size is 64.
% Figure \ref{fig:enc_baselines} visualizes the encoding setups for learning semantic-syntax separation under Bert-GPT2 architectures.
% \begin{figure}[ht!]
%     \centering
%     \includegraphics[scale=0.31]{figs/enc_baselines.pdf}
%     \caption{Encoding setups}
%     \label{fig:enc_baselines}
% \end{figure}

\section{More Experimental results} \label{sec:exp_res}
\begin{figure}[ht!]
    \centering
    \includegraphics[scale=0.34]{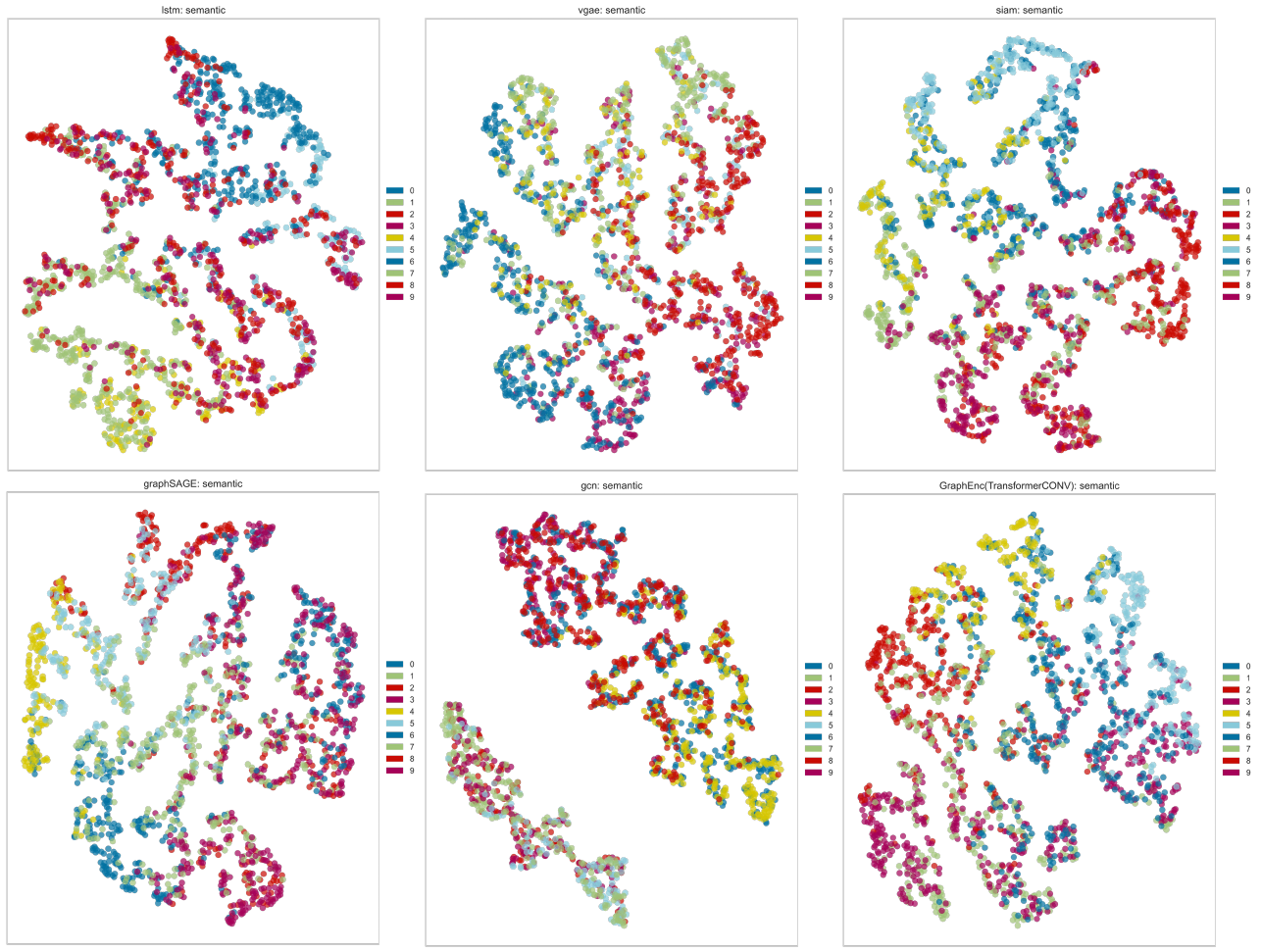}
    \caption{Visualizing semantic space separation (top: LSTM, VGAE, Siam, bottom: graph encoders with GCN, GraphSAGE, TransformerCONV).}
    \label{fig:semantic_space}
\end{figure}

\begin{figure}[ht!]
    \centering
    \includegraphics[scale=0.32]{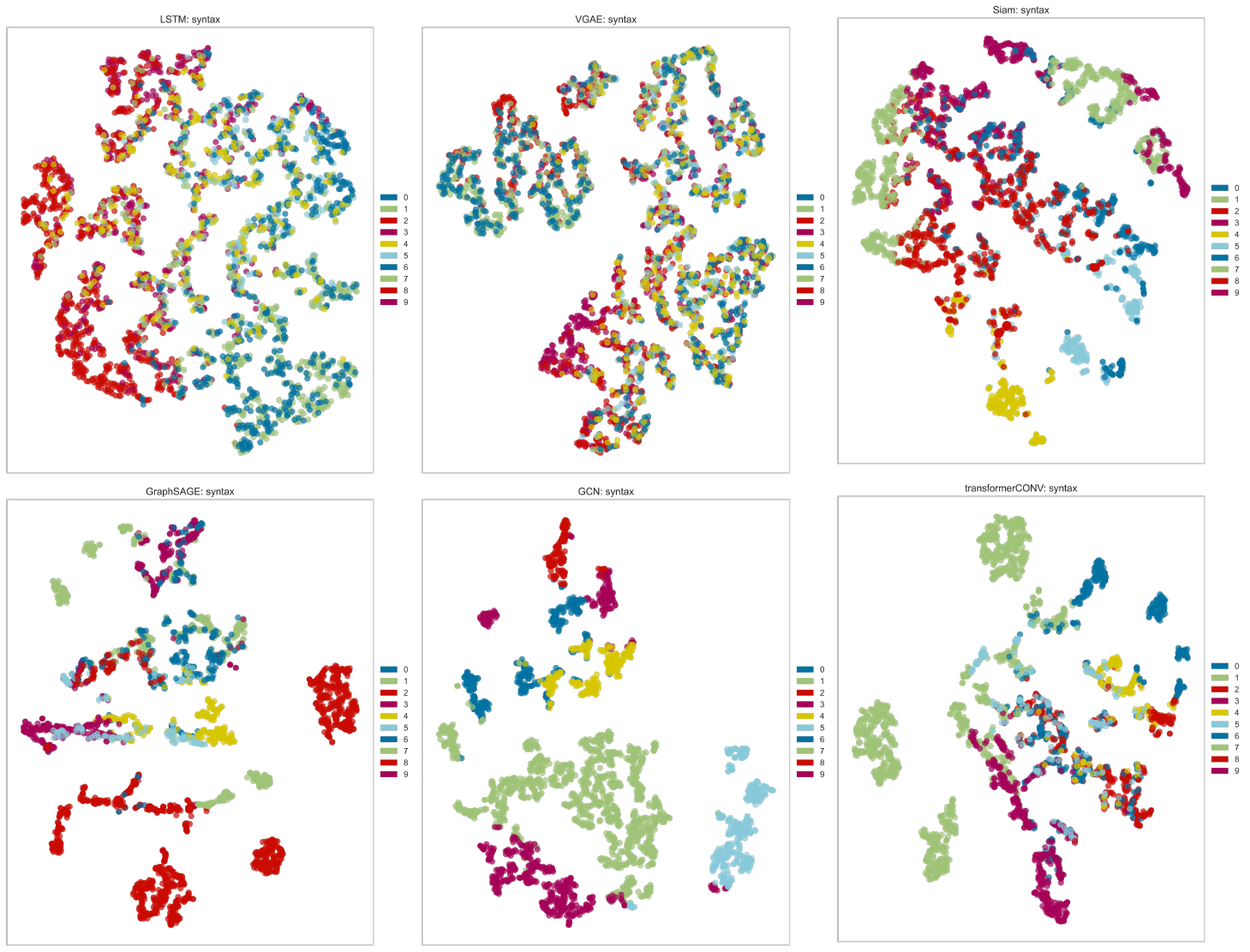}
    \caption{Visualizing syntax space separation (top: LSTM, VGAE, Siam, bottom: graph encoders with GCN, GraphSAGE, TransformerCONV).}
    \label{fig:nl_syntax_space}
\end{figure}

\begin{figure}[ht!]
    \centering
    \includegraphics[scale=0.3]{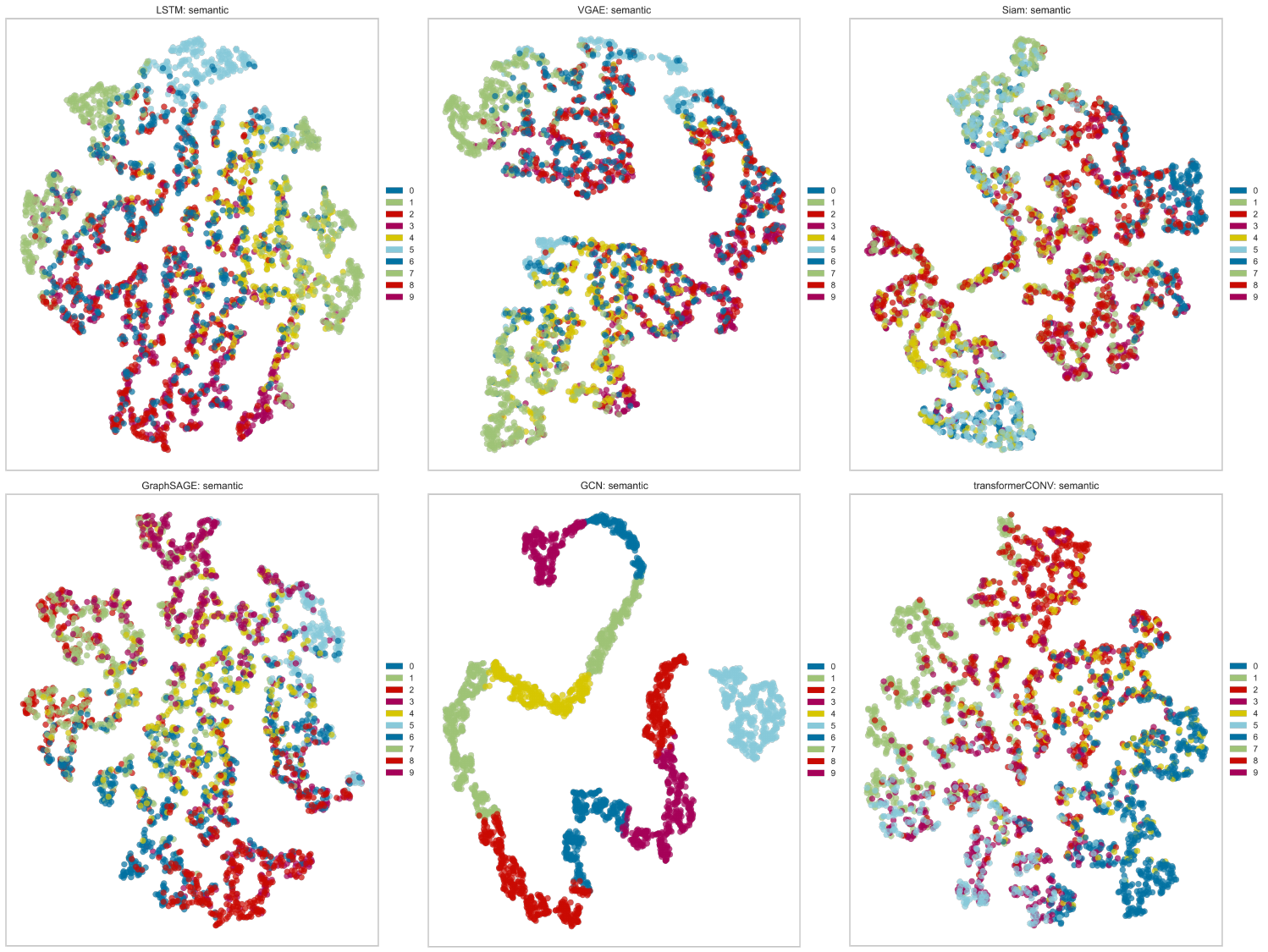}
    \caption{Visualizing semantic space separation (top: LSTM, VGAE, Siam, bottom: graph encoders with GCN, GraphSAGE, TransformerCONV).}
    \label{fig:nl_semantic_space}
\end{figure}

\paragraph{Math Semantic visualization} Figure \ref{fig:semantic_space} visualize the latent space geometry of semantic space.
\paragraph{Explanations Syntax visualization} Figure \ref{fig:nl_syntax_space} visualize the latent space geometry of syntax space of explanatory sentences.
\paragraph{Explanations Semantic visualization} Figure \ref{fig:nl_semantic_space} visualize the latent space geometry of semantic space of explanatory sentences.
\paragraph{Qualitative evaluation} Moreover, we randomly sample the points in each k-mean cluster and output the corresponding sentences or syntax parse tree in Table \ref{tab:trav_examples}, \ref{tab:explanation_gcn_sem_trav_examples}, and \ref{tab:explanation_gcn_syn_trav_examples}.
\begin{table}[ht!]
\begin{tcolorbox}[fontupper=\small, fontlower=\small, middle=0.3cm, top=1pt, bottom=1pt, title=Math symbol: Syntax Cluster Traversal]
$C_0$: Pow(cos(Symbol(E)), Symbol(b)) \\
$C_0$: Pow(exp(Symbol(b)), Symbol(A)) \\
$C_0$: Mul(Symbol(F), sin(Symbol(g))) \\
 
$C_4$: exp(Mul(Pow(Symbol(V), Integer(-1)), Symbol(q))) \\
$C_4$: cos(Mul(Pow(Symbol(b), Integer(-1)), Symbol(g))) \\
$C_4$: exp(Mul(Pow(Symbol(T), Integer(-1)), Symbol(a))) \\
% $C_4$: log(Mul(Pow(Symbol(W), Integer(-1)), Symbol(p)))

$C_8$: sin(Mul(Symbol(A), Symbol(k))) \\
$C_8$: cos(Mul(Symbol(U), Symbol(w))) \\
$C_8$: exp(Mul(Symbol(J), Symbol(l)))
\end{tcolorbox}
\caption{Qualitative evaluation of syntax cluster of Bert-TransCONV encoding.}
\label{tab:trav_examples}
\end{table}
\begin{table}[ht!]
\begin{tcolorbox}[fontupper=\small, fontlower=\small, middle=0.3cm, top=1pt, bottom=1pt, title=Explanations: Semantic Cluster Traversal]
$C_0$: if a pot is exposed to a stove then the pot will become hot \\
$C_0$: if something is used for something else then that something else is the job of that something  \\
$C_0$: if there is a crack in a rock then water can get into the crack  \\
 
$C_8$: decaying plant is a source of nutrients in soil \\
$C_8$: producers are a source of food energy for living things \\
$C_8$: organic matter is a source of nutrients in soil  \\
% $C_4$: log(Mul(Pow(Symbol(W), Integer(-1)), Symbol(p)))

$C_5$: a magnet is a kind of object \\
$C_5$: a board is a kind of object \\
$C_5$: a wagon is a kind of object
\end{tcolorbox}
\caption{Qualitative evaluation of semantic cluster of Bert-GCN encoding.}
\label{tab:explanation_gcn_sem_trav_examples}
\end{table}
\begin{table*}[ht!]
\begin{tcolorbox}[fontupper=\small, fontlower=\small, middle=0.3cm, top=1pt, bottom=1pt, title=Explanations: Syntax Cluster Traversal]
$C_5$: (S (NP (JJ ) (NN )) (VP (VBZ ) (NP (JJ ) (NN )))) \\
$C_5$: (S (NP (DT ) (NN )) (VP (VBZ ) (NP (DT ) (NN ))))\\
$C_5$: (S (NP (JJ ) (JJ ) (NN )) (VP (VBZ ) (NP (JJ ) (NN ))))  \\
 
$C_6$: (S (NP (NN )) (VP (VBZ ) (PP (IN ) (NP (NP (DT ) (NN )) (SBAR (WHNP (WDT )) (S (VP (VBZ ) (VP (VBN ) (PP (IN ) (NP (NN )))))))))))  \\
$C_6$: (S (NP (NN )) (VP (VBZ ) (NP (NP (DT ) (NN )) (PP (IN ) (SBAR (WHADVP (WRB )) (S (NP (DT ) (NN )) (VP (VBZ ) (VP (VBN ))))))))) \\
$C_6$: (S (NP (NN )) (VP (VBZ ) (NP (NP (DT ) (NN )) (SBAR (WHNP (WDT )) (S (VP (VBZ ) (ADJP (JJ ) (JJS ) (PP (IN ) (NP (DT ) (NNP ))))))))))   \\
% $C_4$: log(Mul(Pow(Symbol(W), Integer(-1)), Symbol(p)))

$C_9$: (S (NP (NNS )) (VP (VBP ) (NP (NN )) (PP (IN ) (NP (NNS ))))) \\
$C_9$: (S (NP (NNS )) (VP (VBP ) (PP (IN ) (NP (NN )))))  \\
$C_9$: (S (NP (NNS )) (VP (MD ) (VP (VB ) (NP (NN ) (NN )) (PP (IN ) (NP (DT ) (NN )))))) 
\end{tcolorbox}
\caption{Qualitative evaluation of semantic cluster of Bert-GCN encoding.}
\label{tab:explanation_gcn_syn_trav_examples}
\end{table*}

Besides, in Table \ref{tab:rec}, we provide the comparison of reconstructed sentences between normal Optimus and Bert-TransCONV(addition QKV).
% \begin{table*}[ht!]
% \scriptsize
% \begin{center}
% \begin{tikzpicture}
% \node (table) [inner sep=.1pt] {

\begin{table*}[ht!]
    % \resizebox{\textwidth}{15mm}{
% \begin{tikzpicture}
    % \scriptsize
    \small
    \centering
\renewcommand\arraystretch{1.3}
    \begin{tabular}{p{5cm}p{5cm}p{5cm}}  \toprule
        \textbf{Gold explanations} & \textbf{BERT-GPT2} & \textbf{Bert/TransCONV-GPT2} \\ \hline
        lenses are a kind of object & frog is a kind of object & lenses are a kind of object \\
        the chemical symbol for helium is he & a substance has a physical shape & the chemical symbol for helium is He \\
        a rose is a kind of plant & a window pane is a kind of surface & a rose is a kind of flower \\
        a body of water contains water & a flood has a large amount of rainfall & a body of water contains water \\
        growing is a kind of process & population is a kind of process & growing is a kind of process \\
        air is a kind of gas & farming is a kind of human & air is a kind of gas \\
        action means activity & feed means use & activity means action \\
        soda water is a kind of carbonated beverage & condensing is a kind of change in temperature & soda water is a kind of carbonated beverage \\
        plasma is a kind of state of matter & black probability is a kind of event & plasma is a kind of state of matter \\
        earth is a kind of celestial object & sun is a kind of light & earth is a kind of celestial object \\
        a bee is a kind of living thing & a frog is a kind of amphibian & a bee is a kind of living thing
        \\ 
        green is a kind of color & deforestation is a kind of process & green is a kind of color \\ 
        a wooded area is a kind of forest & a coal mine is a kind of natural resource & a wooded area is a kind of forest \\ \toprule
    \end{tabular}
    \caption{Explanation reconstruction (left: original explanations from WorldTree corpus, middle: explanations from Optimus, right: explanations from Bert-TransCONV (addition Q)).} 
    \label{tab:rec}
\end{table*}

\begin{figure*}[ht!]
    \centering
    \includegraphics[scale=0.4]{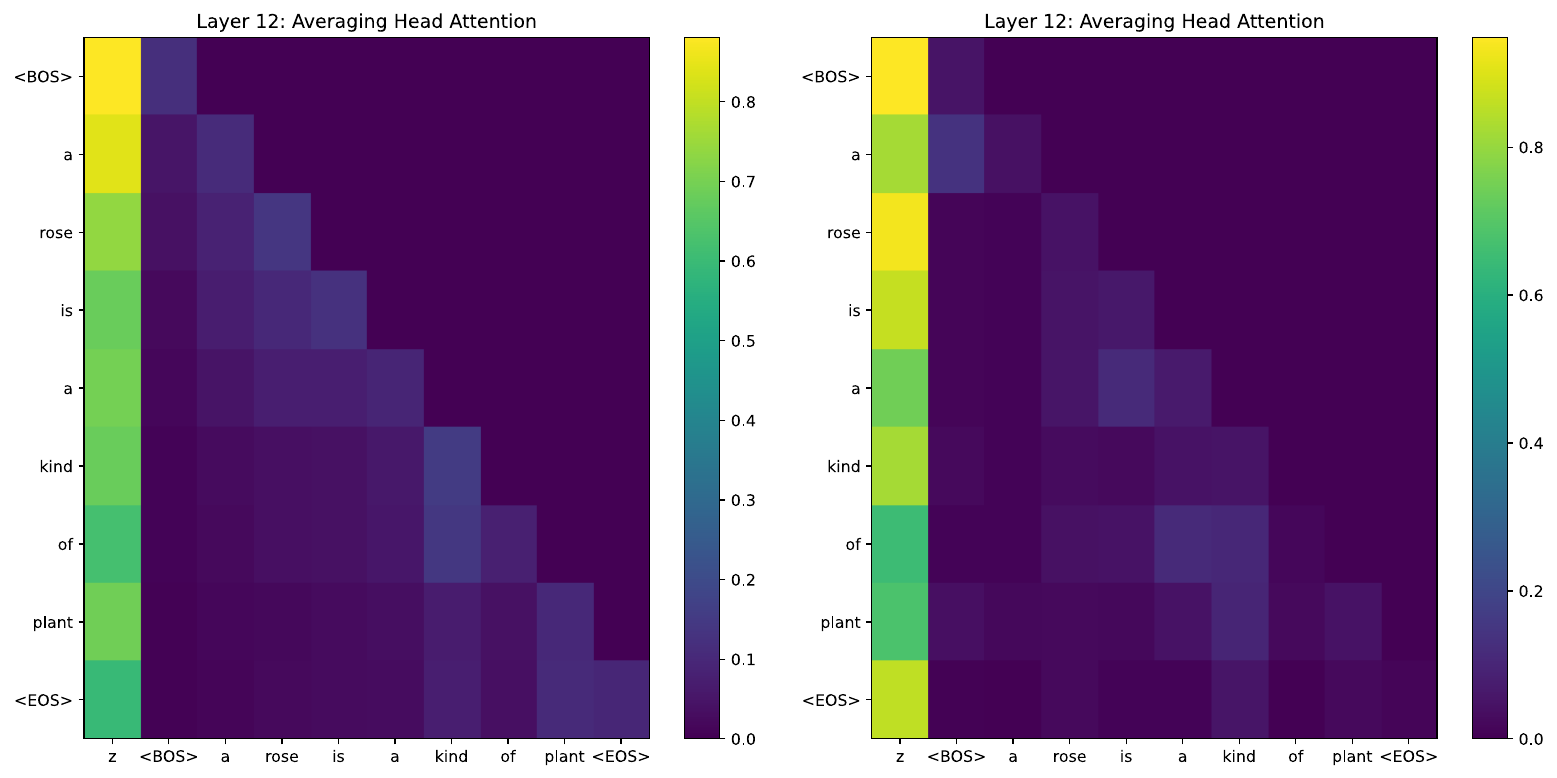}
    \caption{\textit{a rose is a kind of plant}.}
    \label{fig:heatmap_1}
\end{figure*}

\begin{figure*}[ht!]
    \centering
    \includegraphics[scale=0.35]{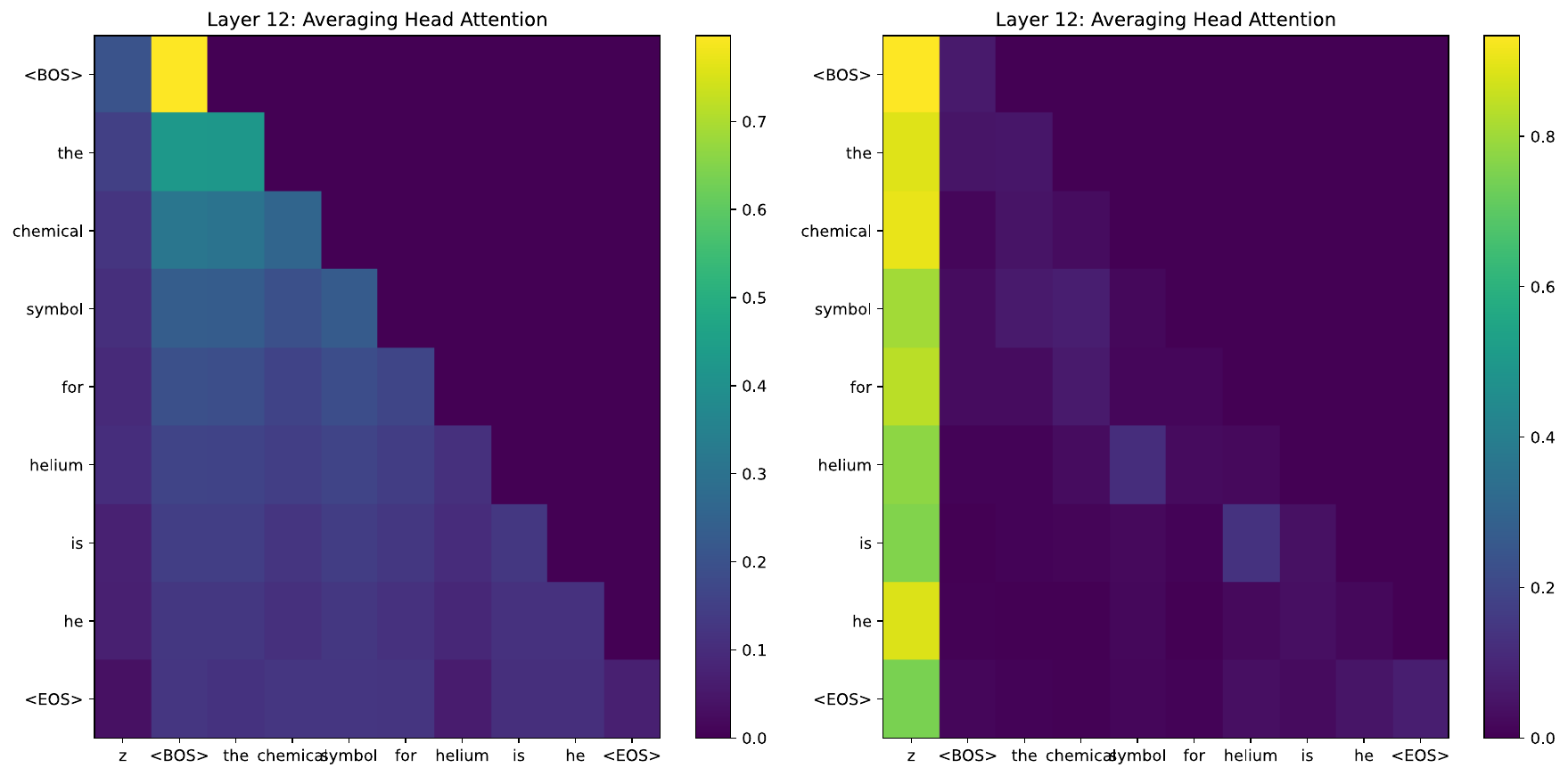}
    \caption{\textit{the chemical symbol for helium is he}.}
    \label{fig:heatmap_2}
\end{figure*}

\paragraph{Attention heatmap} \label{sec:attn_map} We provide more attention heatmap of different sentences in Figure \ref{fig:heatmap_1} and \ref{fig:heatmap_2}. Similar observation as before, the latent representation can better capture word content information under the graph-language encoding setup.

\paragraph{Traversal} \label{sec:app_traversal} We provide the traversed sentences in table \ref{tab:traversal_example}. From it, we can observe that the semantic-syntax separation can better hold the syntax structures of the traversed sentences and have the potential to hold similar semantics, indicating the graph-induced latent semantic geometry is more regular than that of normal Optimus, resulting in better controllability during decoding. We also provide the traversed examples of syntax space in table \ref{tab:traversal_example_1}. From it, we can observe that the generated sentences hold similar semantics about \textit{sea} and \textit{water} as input, compared to normal Optimus which generates unrelated semantics: \textit{desert} and \textit{forest}, etc.
\begin{table*}[t]
\begin{tcolorbox}[fontupper=\small, fontlower=\small, middle=0.3cm, top=1pt, bottom=1pt, title=Semantic Space Traversal]
Input: \textit{a sea is a source of sea water} \\
 0: a desert is a land found in desert environments \\
 1: a forest is a large structure that contains lots of trees \\
 2: a river is a nonliving thing \\
 3: a canyon is a very deep valley \\
 4: a mountain is a large land mass
    \\ \\
 0: a sea is a source of water for humans \\
 1: a sea is a source of freshwater \\
 2: a river is a source of water \\
 3: an ocean is a source of water for residents
\end{tcolorbox}
\caption{Qualitative evaluation of traversed examples of Optimus (top) and Bert-TransCONV (addition QKV) (bottom).}
\label{tab:traversal_example}
\end{table*}
\begin{table*}[t]
\begin{tcolorbox}[fontupper=\small, fontlower=\small, middle=0.3cm, top=1pt, bottom=1pt, title=Syntax Space Traversal]
Input: \textit{a sea is a source of sea water} \\
 0: a river is synonymous with a coastline \\
 1: a hurricane is composed of water vapor and dust \\
 2: a hurricane is the source of most of water vapor in the atmosphere \\
 3:  hurricane is mainly made of water vapor \\
 4: a hurricane is measuring the amount of water in an area
\end{tcolorbox}
\caption{Qualitative evaluation of traversed examples of Bert-TransCONV (addition QKV).}
\label{tab:traversal_example_1}
\end{table*}

\end{document}